%% file: iclr2027_conference.tex
\documentclass{article} % For LaTeX2e
\usepackage{iclr2027_conference,times}

\input{math_commands.tex}

\usepackage[utf8]{inputenc} % allow utf-8 input
\usepackage[T1]{fontenc} % use 8-bit T1 fonts
\usepackage{hyperref} % hyperlinks
\hypersetup{colorlinks=true,linkcolor=cyan,citecolor=cyan,urlcolor=cyan}
\usepackage{url} % simple URL typesetting
\usepackage{booktabs} % professional-quality tables
\usepackage{array} % vertically centered paragraph columns
\usepackage{amsmath}
\usepackage{amssymb}
\usepackage{amsfonts} % blackboard math symbols
\usepackage{amsthm}
\usepackage{graphicx}
\usepackage{nicefrac} % compact symbols for 1/2, etc.
\usepackage{microtype} % microtypography
\usepackage{xcolor} % colors
\usepackage{colortbl} % subtle row shading in tables
\usepackage{multirow}
\usepackage{wrapfig}
\usepackage{bm}

\makeatletter
\newcommand{\appendixlocaltoc}{%
  \begingroup
  \hypersetup{linkcolor=black}%
  \setcounter{tocdepth}{2}%
  \section*{Appendix Contents}%
  \@starttoc{atoc}%
  \endgroup
}
\newcommand{\enableappendixlocaltoc}{%
  \let\appendix@original@addcontentsline\addcontentsline
  \renewcommand{\addcontentsline}[3]{%
    \appendix@original@addcontentsline{##1}{##2}{##3}%
    \def\appendix@target{##1}%
    \def\appendix@toc{toc}%
    \ifx\appendix@target\appendix@toc
      \appendix@original@addcontentsline{atoc}{##2}{##3}%
    \fi
  }%
}
\makeatother

\title{Inertial Video World Models}

\author{Yuan Gao$^{1}$\thanks{Equal contribution. $^{ \dagger}$Corresponding author.} , Fan Xu$^{2 *}$, Yu Guo$^{3}$, Yuxu Lu$^{4}$, \\\textbf{Zhengru Fang$^{5}$, Hao Wu$^{1\dagger}$, Xiaomeng Huang$^{1\dagger}$} \\
$^1$Tsinghua University, $^2$University of Science and Technology of China, \\$^3$City University of Hong Kong, $^4$Hong Kong Polytechnic University, \\$^5$The Hong Kong University of Science and Technology
}

\iclrfinalcopy % Uncomment for camera-ready version, but NOT for submission.
\begin{document}

\newcommand{\model}{\textsc{InertialWorld}}
\newcommand{\method}{\model}

\maketitle

\input{sec/0_abstract}
\input{sec/1_intro}
\input{sec/2_Related_work}
\input{sec/3_Method}
\input{sec/4_Experiments}

\input{sec/5_conclusions}

\clearpage

\subsection*{AI use statement}

We use AI to check for grammatical errors in the text; the AI is not involved in ideas, experimental design, or the actual writing of the paper.

% \subsection*{Ethics statement}

% This work complies with the ICLR Code of Ethics. Human evaluations were conducted with informed consent from volunteers. All datasets and pretrained models are publicly available and used under their respective licenses. We have carefully designed our experiments to minimize risks of bias or discrimination. We uphold the principles of transparency and integrity throughout the work.

\subsection*{Ethics statement}

Human evaluations were conducted with informed consent from volunteers. All datasets and pretrained models are publicly available and used under their respective licenses. We have carefully designed our experiments to minimize risks of bias or discrimination. We uphold the principles of transparency and integrity throughout the work.

\subsection*{Reproducibility Statement}
The method section specifies the causal interface, recurrent video state, memory
transition and readout, and training objective. The implementation
records model configurations, fixed data splits, checkpoint-selection rules,
and the sampling and optimizer states required to resume training.

\section*{Acknowledgements}
This work was supported by the National Natural Science Foundation of China (42125503, 42430602).

\bibliography{iclr2027_conference}
\bibliographystyle{iclr2027_conference}

\clearpage
\appendix
\enableappendixlocaltoc
\appendixlocaltoc
\clearpage
% Keep appendix float spacing compact and prevent \flushbottom from stretching
% the gap after a table when the remainder of a page is lightly populated.
\raggedbottom
\setlength{\textfloatsep}{10pt plus 2pt minus 2pt}
\setlength{\floatsep}{8pt plus 2pt minus 2pt}
\setlength{\intextsep}{10pt plus 2pt minus 2pt}
\input{sec/X_suppl}

\end{document}

%% file: math_commands.tex
\usepackage{amsmath,amsfonts,bm}

\def\eqref#1{equation~\ref{#1}}
\def\plaineqref#1{\ref{#1}}
\def\1{\bm{1}}

\DeclareMathAlphabet{\mathsfit}{\encodingdefault}{\sfdefault}{m}{sl}
\SetMathAlphabet{\mathsfit}{bold}{\encodingdefault}{\sfdefault}{bx}{n}

%% file: sec/0_abstract.tex
\begin{abstract}
The physical world has inertia: it evolves from what already exists rather than being recreated at every instant. However, many existing video world models lack an explicit mechanism for inertial state evolution, leaving the continuity of motion and structure implicit in video prediction. During autoregressive rollouts, repeated reconstruction can accumulate errors and destabilize the predicted dynamics. To address this limitation, we introduce \method{}, an Inertial Video World Model that makes inertia a structural prior for autoregressive prediction: existing scene evidence is carried across time rather than reconstructed at every step. Central to this formulation is Inertial Memory, which turns memory from passive predictive context into an evolving video state with two complementary views: a dense Eulerian grid preserves scene-aligned evidence, while persistent Lagrangian carriers bind local content to learned canonical coordinates. At each step, the model transports the carrier content according to action-conditioned motion, predicts the complete next video state from the transported carriers and dense grid, and uses a learned gate to combine transported and newly predicted content into the memory for the next step. Extensive experiments across multiple domains, including robot manipulation, 3D game simulation, and open-world navigation, demonstrate that \method{} achieves competitive autoregressive prediction performance against strong video world model baselines. Codes link:~\url{https://github.com/YuanGao-YG/InertialWorld}.
\end{abstract}

%% file: sec/1_intro.tex
\section{Introduction}
\label{sec:introduction}

Learned world models provide predictive environments for planning and control by modeling action-conditioned dynamics \citep{ha2018world,hafner2019learning,hafner2025mastering}. Within this broader family, video world models instantiate learned dynamics through action-conditioned video prediction, enabling interactive simulation in games, robotic manipulation, and embodied navigation \citep{micheli2022transformers,bruce2024genie,yang2023learning,bar2025navigation}. Autoregressive and diffusion-based formulations have substantially improved generation fidelity and controllability \citep{wu2024ivideogpt,alonso2024diamond,huang2025vid2world,zhu2026astra}, while post-training further refines predicted rollouts \citep{wu2026rlvr,wu2026roboalign}. Despite these advances, physical evolution is inertial: the world evolves from what already exists rather than being recreated at every instant. Most video world models, however, still formulate temporal evolution as repeated conditional generation, predicting persistent scene content again at every step rather than explicitly carrying it forward. Since each generated state conditions the next transition, small reconstruction errors can repeatedly re-enter the rollout and destabilize predicted motion and structure \citep{wang2017predrnn,babaeizadeh2021fitvid}.

A prominent line of research mitigates this limitation by augmenting video prediction with memory. Recurrent video predictors retain temporal features across prediction steps \citep{wang2017predrnn,wang2018predrnn++}, while transformer world models expose longer histories to the transition function \citep{chen2022transdreamer,wu2024ivideogpt}. This additional context is particularly valuable when occlusion or accumulated generation noise makes the current frame ambiguous. Earlier features can retain identity and layout cues that are no longer directly visible but remain relevant to the next transition. Spatial and persistent memories preserve scene-aligned evidence across recurrent prediction steps \citep{wu2026video,zhou2026learning}. Object-centric representations provide a complementary form of persistence by binding content to structured entities \citep{locatello2020object,kipf2021conditional}, and their video extensions model the dynamics of these entities through time \citep{wu2022slotformer}. Collectively, these mechanisms expose historical evidence to subsequent state transitions and improve temporal consistency, but retention alone does not specify how the retained content should evolve. When memory serves as context for a generic transition, the transition remains free to synthesize the next state without explicitly advancing the stored content. Transformation-based predictors demonstrate the value of warping observations or features \citep{jaderberg2015spatial,finn2016unsupervised}, yet warping does not by itself make transport the recurrent evolution rule of persistent memory across model-predicted steps. Thus, memory determines which past evidence remains available, whereas inertia additionally specifies what persists and where it moves. This distinction motivates the central question of this work: \textbf{\emph{How can a video world model evolve the world it already remembers, rather than repeatedly reconstructing its persistent content?}}

\begin{figure}[t]
  \centering
  \includegraphics[width=\linewidth]{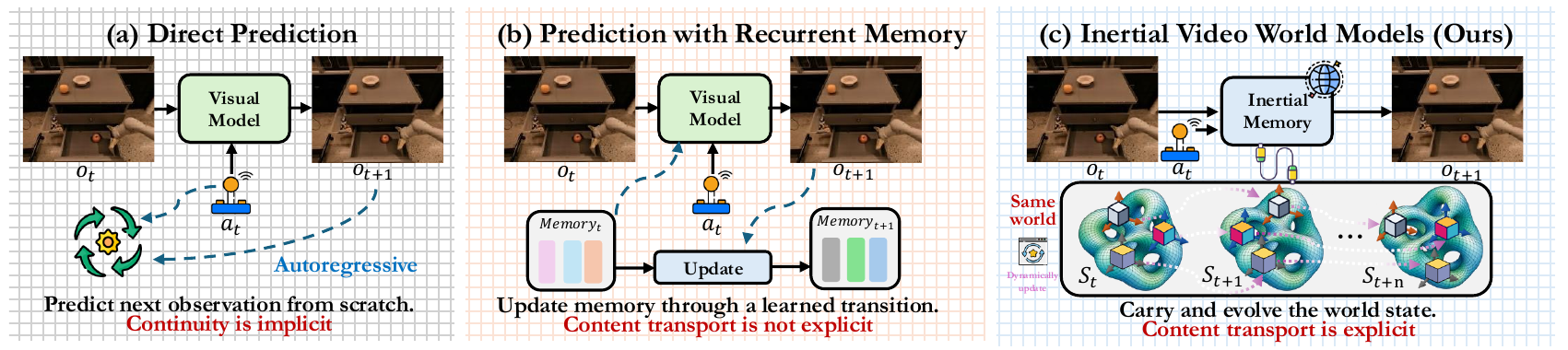}
  \vspace{-20pt}
  \caption{\textbf{From memory to inertia.} All three paradigms predict $\mathbf{o}_{t+1}$ from $\mathbf{o}_t$ and action $\mathbf{a}_t$. (a) Direct prediction reconstructs each next observation, leaving persistence implicit. (b) Recurrent memory retains the past as predictive context, but a generic transition need not use predicted motion to advance stored content. (c) \method{} actively evolves its memory: action-conditioned coordinate motion carries bound content before gated prediction updates the state.}
  \vspace{-15pt}
  \label{fig:intro_overview}
\end{figure}

To address this question, we introduce \method{}, an Inertial Video World Model that makes inertia a structural prior for autoregressive prediction: existing scene evidence is carried across time rather than reconstructed at every step. As illustrated in Figure~\ref{fig:intro_overview}, its central component, \emph{Inertial Memory}, turns retained predictive context into an evolving video state with two complementary views: a dense Eulerian grid preserves scene-aligned evidence, while persistent Lagrangian carriers bind local content to learned canonical coordinates. At each step, the model transports the carrier content according to action-conditioned motion, predicts the complete next video state from the transported carriers and dense grid, and uses a learned gate to combine transported and newly predicted content into the memory for the next step. This carry-before-write transition establishes persistence as the default path of state evolution while preserving the capacity to model changes that transport alone cannot explain, without requiring object annotations or calibrated physical parameters. We evaluate \method{} on RT-1 robot manipulation \citep{brohan2022rt}, Counter-Strike 3D game simulation \citep{huang2025vid2world}, and RECON open-world navigation \citep{shah2021rapid}. Across all three interactive domains, \method{} produces competitive autoregressive predictions against strong video world model baselines. Our contributions are threefold:
\begin{itemize}
  \item We formulate \emph{inertial video world modeling}, a state-evolution view of video world models in which existing scene evidence is inherited by default rather than repeatedly reconstructed.

  \item We introduce \emph{Inertial Memory}, a hybrid Eulerian--Lagrangian video state that preserves dense scene evidence while transporting persistent regional content through action-conditioned coordinate motion before a learned gate updates the memory.

  \item We conduct extensive autoregressive prediction evaluations across robot manipulation, 3D game simulation, and open-world navigation, demonstrating competitive autoregressive prediction across all three domains against strong video world model baselines.
\end{itemize}

%% file: sec/2_Related_work.tex
\section{Related Work}
\label{sec:related_work}

\noindent\textbf{Video World Models.}
Action-conditioned video prediction learns how future observations evolve under control \citep{oh2015action,finn2016unsupervised}. Early approaches use stochastic, recurrent, and physics-inspired dynamics to model uncertainty and temporal dependence \citep{denton2018stochastic,wang2017predrnn,guen2020disentangling,babaeizadeh2021fitvid}. Learned discrete representations \citep{van2017neural,esser2021taming} enable scalable autoregressive and masked video prediction \citep{yan2021videogpt,yu2023magvit,yu2024language,wang2024omnitokenizer}. These formulations support interactive simulation across games and robotics \citep{bruce2024genie,valevski2025diffusion,wu2024ivideogpt,yang2023learning, zhou2024dino,zhou2024robodreamer}, while diffusion-based models provide an alternative generative backbone \citep{alonso2024diamond,huang2025vid2world}. Reinforcement learning and alignment objectives further refine predicted rollouts \citep{wu2026rlvr,wu2026roboalign}. In parallel, latent world models learn compact predictive states for planning and imagined policy learning \citep{hafner2019learning,hafner2020mastering,hafner2025mastering}, sometimes prioritizing task-relevant dynamics over frame reconstruction \citep{schrittwieser2020mastering,hansen2024tdmpc2}. Rather than changing the video generator or its training objective, our work changes how the recurrent video state evolves by introducing transport-structured persistence that is compatible with different generative backbones.

\noindent\textbf{Memory and Structured Video State.}
Existing memory mechanisms primarily determine \emph{what} a video world model retains. Recurrent states and token histories preserve temporal context \citep{wang2017predrnn,wang2018predrnn++}; spatial and persistent memories preserve scene-aligned content \citep{wu2026video,zhou2026learning}; and object-centric representations bind content to persistent entities \citep{burgess2019monet,greff2019multi,locatello2020object}, with temporal extensions supporting tracking and prediction \citep{kipf2021conditional,wu2022slotformer,kipf2019contrastive}. Relational models expose interactions between state elements \citep{battaglia2016interaction,kipf2018neural,battaglia2018relational}, while Hamiltonian and graph-based simulators impose additional structure on state transitions \citep{greydanus2019hamiltonian,sanchez2020learning}. These approaches enrich what persists, but they do not by themselves specify \emph{how} retained content should evolve. A generic recurrent update can encode motion without requiring that motion to relocate content already in memory. Transformation-based predictors demonstrate the value of warping observations or features \citep{finn2016unsupervised,jaderberg2015spatial}, but warping alone does not make transport the recurrent evolution rule of a persistent state. Inertial Memory instead treats persistent canonical evidence as the transported quantity: it maintains a dense scene field, binds selected local content to learned coordinate frames, and uses their predicted evolution to transport the bound content before the memory is revised. Its defining property is therefore the explicit coupling of coordinate evolution and content transport within a hybrid dense--regional video state, rather than recurrence, slots, or warping in isolation.

%% file: sec/3_Method.tex
\section{Method}
\label{sec:method}

\subsection{Problem Definition}
\label{sec:problem}

\begin{figure}[t]
  \centering
\includegraphics[pagebox=cropbox,width=\linewidth]{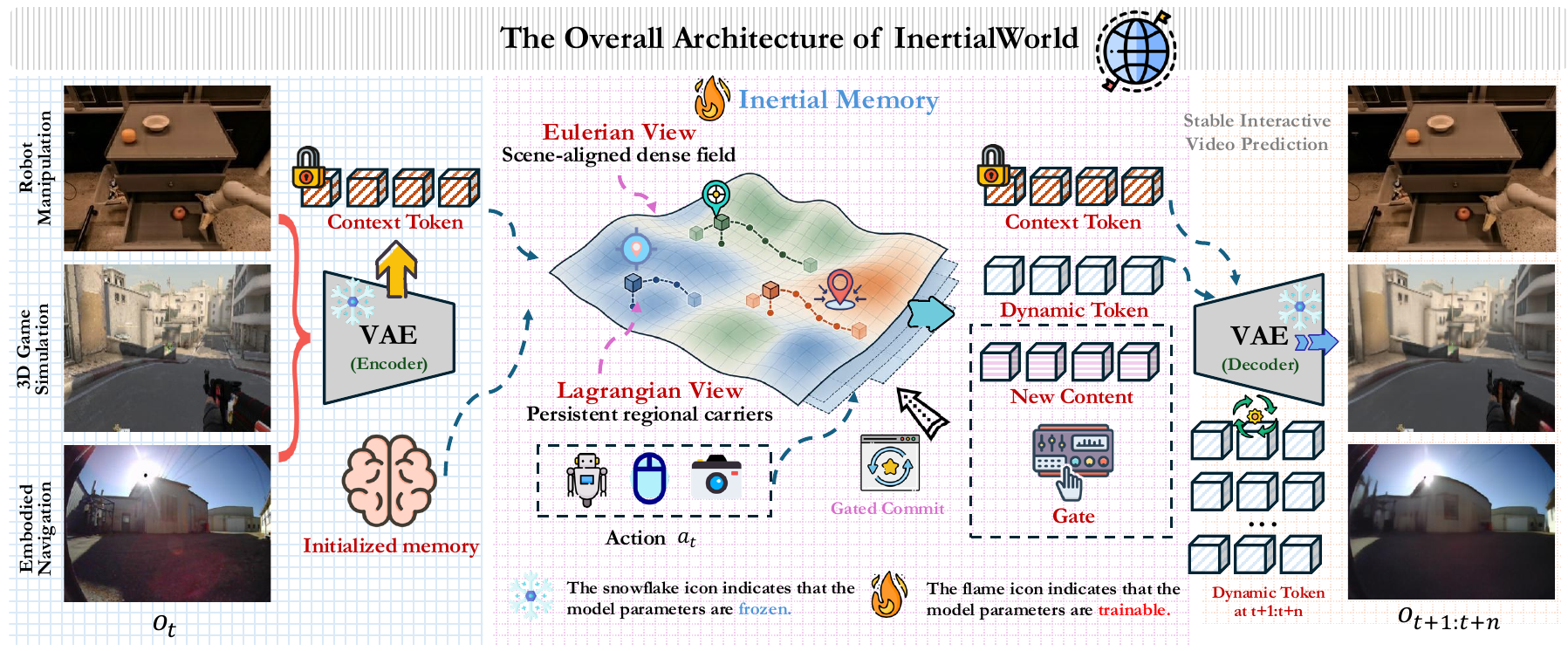}
\vspace{-20pt}
  \caption{\textbf{Overall architecture of \method{}.} A frozen encoder extracts the rollout-level Context code and initializes Inertial Memory. Its dense Eulerian grid preserves scene-aligned evidence, while action-conditioned Lagrangian carriers transport persistent regional content. A learned gate combines carried and newly predicted content, and a frozen decoder reconstructs future frames from the fixed Context code and predicted Dynamic codes.}
  \label{fig:main_architecture}
  \vspace{-18pt}
\end{figure}

Figure~\ref{fig:main_architecture} summarizes the complete prediction pipeline. At time $t$, a video world model receives the current frame $\mathbf{o}_t$ and action $\mathbf{a}_t$ and predicts $\widehat{\mathbf{o}}_{t+1}$. A multi-step rollout observes $\mathbf{o}_t$ only once; all subsequent frames are generated from the model's own state and the action sequence. We use a frozen codec that represents the observed scene with a rollout-level Context code $\mathbf{C}$ and each predicted frame with a Dynamic code $\mathbf{D}$. We encode $\mathbf{C}=\operatorname{Enc}_{C}(\mathbf{o}_t)$ and initialize the recurrent video state as $\mathcal{S}_t=\operatorname{Init}(\mathbf{C})$. Each causal prediction step is then
\begin{equation}
  (\widehat{\mathbf{D}}_{\tau+1},\mathcal{S}_{\tau+1})
  =\mathcal{F}_{\theta}(\mathcal{S}_{\tau},\mathbf{a}_{\tau};\mathbf{C}),
  \qquad
  \widehat{\mathbf{o}}_{\tau+1}
  =\operatorname{Dec}(\mathbf{C},\widehat{\mathbf{D}}_{\tau+1}).
  \label{eq:causal_step}
\end{equation}
The Context code remains fixed and anchors the observed scene throughout the rollout. The predicted Dynamic code is a complete code for the next frame, not a residual correction to a ground-truth frame. At the first transition, there is no previous Dynamic code; all observed information enters through $\mathbf{C}$ and the initialized state. Future frames define training targets $\mathbf{D}^{\star}_{\tau+1}$, but neither those targets nor their encoded features are fed back into the recurrence. This causal interface defines the prediction problem but does not specify how the state should evolve. A generic recurrent state can retain past information while the transition still rewrites its content at every step. We instead make persistence structural: predicted motion first acts on stored content, and prediction updates the state only after this content has been carried forward.

\subsection{Inertial Memory}
\label{sec:inertial_memory}

Inertial Memory realizes this carry-before-write principle by coupling content with learned coordinates. Its state is $\mathcal{S}_{\tau}=(\mathbf{G}_{\tau}, \{\mathbf{r}_{\tau,k}\}_{k=1}^{K},\mathbf{B}_{\tau})$, where $\mathbf{G}_{\tau}$ is a dense Eulerian grid and $\mathbf{r}_{\tau,k}=(\mathbf{A}_{\tau,k}, \boldsymbol{\alpha}_{\tau,k},\boldsymbol{\xi}_{\tau,k})$ is a persistent Lagrangian carrier. The grid remains aligned with scene coordinates and preserves evidence across the full spatial domain. Each carrier instead moves with a local region: $\mathbf{A}_{\tau,k}$ stores canonical content, $\boldsymbol{\alpha}_{\tau,k}$ describes its support, and $\boldsymbol{\xi}_{\tau,k}$ specifies where that content is placed. The soft binding map $\mathbf{B}_{\tau}$ assigns evidence between the carriers and a dense fallback. The carriers are learned regions rather than supervised semantic objects, so this representation does not require object annotations or a complete object decomposition.

\paragraph{Carrying persistent content.}
Let $\boldsymbol{\xi}_{\tau}$ collect all carrier coordinates. The first part of a transition predicts their action-conditioned motion and applies it to the stored state:
\begin{equation}
  \boldsymbol{\xi}_{\tau+1}
  =U_{\theta}(\mathcal{S}_{\tau},\mathbf{a}_{\tau}),
  \qquad
  \widetilde{\mathcal{S}}_{\tau+1}
  =\operatorname{Carry}(\mathcal{S}_{\tau};
  \boldsymbol{\xi}_{\tau}\!\rightarrow\!\boldsymbol{\xi}_{\tau+1}).
  \label{eq:carry}
\end{equation}
The coordinate update $U_{\theta}$ predicts bounded motion from current memory and action. Carry re-renders each carrier's canonical content and support at its new coordinates, so local evidence moves with the carrier rather than being synthesized again at the destination. For the dense grid, the binding map combines carrier-induced motion with a learned fallback flow; normalized forward splatting propagates evidence while accounting for ownership and visibility. The resulting $\widetilde{\mathcal{S}}_{\tau+1}$ is therefore the state inherited from the past. When the predicted motion is zero, this operation reduces to an identity carry. The coordinates are learned representational variables, not calibrated physical parameters.

\paragraph{Predicting and updating.}
Transport alone cannot explain newly visible content, deformation, or other changes not represented by the current memory. Let $\widetilde{\mathcal{M}}_{\tau+1}$ denote the writable content in the carried state. A predictor reads that state and jointly produces the complete next Dynamic code, a candidate memory update, and a learned write gate. Suppressing time subscripts within this transition and using $+$ for time $\tau+1$, this is
\begin{equation}
  (\widehat{\mathbf{D}}^{+},\mathcal{M}^{\mathrm{new}},\mathbf{g})
  =P_{\theta}(\widetilde{\mathcal{S}},\mathbf{a};\mathbf{C}),
  \qquad
  \mathcal{M}^{+}=(1-\mathbf{g})\odot\widetilde{\mathcal{M}}
  +\mathbf{g}\odot\mathcal{M}^{\mathrm{new}}.
  \label{eq:predict_commit}
\end{equation}
The first output is decoded into the complete next frame, allowing the model to generate changes that cannot be obtained by transport alone. The second output proposes new writable memory content, while the gate determines which carried values should persist and which should be replaced. Together with the transported carrier coordinates, the updated content forms $\mathcal{S}_{\tau+1}$. Thus, persistence is the default path, but the memory can still adapt when the newly predicted state provides evidence for change. Because the proposal is generated from the model prediction rather than the target Dynamic code, repeated application of Eqs.~\plaineqref{eq:carry}--\plaineqref{eq:predict_commit} remains strictly predicted-only. This ordered transition---carry, predict, then update---is our operational definition of representational inertia; it does not impose a hard-coded Newtonian law.

\subsection{Learning and Autoregressive Rollout}
\label{sec:objective}

We train on the same predicted states used at inference and backpropagate through the complete rollout. Discrete internal decisions remain hard on the forward path and use a straight-through estimator for optimization. The recurrence is never refreshed with a target state or detached between autoregressive steps. We progressively increase the prediction horizon from $8$ to $16$ and then $32$ steps so that the model learns to preserve and update its own state under repeated application.

The training objective combines prediction, memory-consistency, and regularization terms:
\begin{equation}
  \mathcal{L}=\mathcal{L}_{\mathrm{pred}}+\lambda_{\mathrm{mem}}\mathcal{L}_{\mathrm{mem}}+\lambda_{\mathrm{reg}}\mathcal{L}_{\mathrm{reg}}.
  \label{eq:training_objective}
\end{equation}
Here, $\mathcal{L}_{\mathrm{pred}}$ supervises the complete Dynamic token IDs, their finite scalar quantization digits~\citep{mentzer2024finite}, and the decoded frame's appearance and temporal change. $\mathcal{L}_{\mathrm{mem}}$ provides consistency signals for transporting, binding, and updating memory from adjacent frozen-codec representations without object or contact annotations. $\mathcal{L}_{\mathrm{reg}}$ regularizes carrier diversity and dense motion.

We use $\lambda_{\mathrm{mem}}=\lambda_{\mathrm{reg}}=1$ in all experiments; each grouped loss is a fixed weighted sum of its constituent terms. Appendix~\ref{app:wm_architecture} gives the complete constituent list and coefficients together with the full staging schedule and gate initialization.

Ground truth is used only to compute these losses; it never replaces $\mathcal{S}_{\tau}$ in the rollout. Training therefore matches the causal, predicted-only recurrence used at inference.

%% file: sec/4_Experiments.tex
\section{Experiments}
\label{sec:experiments}

Our experiments address four questions that follow directly from the proposed inertial principle: (i) does the complete video world model improve the accuracy and stability of autoregressive prediction across interaction domains, (ii) are its gains retained when evaluation focuses on action-induced motion, (iii) do its generated rollouts preserve behavioral differences between policies, and (iv) does performance degrade when Inertial Memory is removed? We answer these questions in the same order below.

\subsection{Stable Autoregressive Video Prediction}

\noindent\textbf{Evaluation Protocols.}
We first evaluate autoregressive video prediction on RT-1 \citep{brohan2022rt}. Every method is evaluated on the same 1,024 episodes and predicts 32 future frames using its native observation interface. All outputs are compared with the same $256\!\times\!320$ ground-truth frames. We evaluate the released checkpoints of RLVR-World~\citep{wu2026rlvr}, Vid2World~\citep{huang2025vid2world}, WorldGym~\citep{quevedo2026worldgym}, and DriftWorld~\citep{lu2026driftworld} without retraining. The frozen protocol is disjoint from our training and development episodes; because released methods use different native RT-1 partitions, it is not necessarily held out for every baseline. Under the released RLVR-World partition, 1,015 of the 1,024 evaluation episodes belong to its training subset. We therefore treat this as a descriptive released-checkpoint comparison, rather than a controlled estimate of held-out generalization. For Counter-Strike 3D game simulation, we follow the CSGO benchmark and data partition used by Vid2World~\citep{huang2025vid2world}. From its 500 held-out episodes, we reserve the first 128 for validation and evaluate a fixed 70-episode subset of the remaining split. Each selected episode contributes one fixed anchor with at least 32 preceding frames, and every method predicts the same 32 adjacent targets at $256\!\times\!320$ resolution. We compare against the released checkpoints of Vid2World and DIAMOND~\citep{alonso2024diamond}; each method retains its native conditioning interface, while conditioning frames are never scored. Appendix~\ref{app:datasets_protocols} specifies the frozen selection rule and the complete frame--action alignment. We additionally evaluate video prediction for embodied navigation on RECON~\citep{shah2021rapid}. We freeze 128 distinct test episodes from the official 2,367-episode held-out split after excluding the 17 trajectories used for validation. The common anchor in each episode provides the last observed frame $o_t$, and every method predicts the same 32 adjacent targets $o_{t+1:t+32}$ at $240\!\times\!320$ resolution; baselines retain their native four-frame context when required. We compare the released checkpoints of Vid2World, NWM~\citep{bar2025navigation}, and RAE-NWM~\citep{zhang2026rae}. For all three dataset blocks in Table~\ref{tab:main_results}, we first average each frame-level metric over the 32 predicted frames of an episode. We then report the mean and sample standard deviation across 1,024 RT-1 episodes, 70 CSGO test episodes, and 128 RECON test episodes, respectively; conditioning frames are never scored.

\begin{table}[t]
  \centering
  \caption{Autoregressive video prediction on RT-1, CSGO, and RECON using released baseline checkpoints. Every method predicts 32 future frames. Best results within each dataset are in \textbf{bold}, and second-best results are \underline{underlined}. Pixel errors are measured on RGB values normalized to $[0,1]$.}
  \label{tab:main_results}
  \vspace{0.4em}
  \small
  \setlength{\tabcolsep}{5pt}
  \renewcommand{\arraystretch}{1.16}
  \begin{tabular*}{\linewidth}{@{\extracolsep{\fill}}lcccc@{}}
    \toprule
    \rowcolor{black!8}
    \multicolumn{5}{c}{\textit{Robot Manipulation: RT-1}} \\
    \midrule
    Model & PSNR $\uparrow$ & SSIM $\uparrow$ & LPIPS $\downarrow$ & MSE $\downarrow$ \\
    \midrule
    RLVR-World
      & $\underline{21.95 \pm 2.36}$
      & $0.780 \pm 0.062$
      & $\underline{0.168 \pm 0.050}$
      & $\underline{0.0086 \pm 0.0053}$ \\
    \rowcolor{black!4}
    Vid2World
      & $21.32 \pm 2.35$ & $0.724 \pm 0.072$ & $0.244 \pm 0.050$ & $0.0101 \pm 0.0060$ \\
    WorldGym
      & $21.72 \pm 2.45$
      & $\underline{0.782 \pm 0.063}$
      & $0.205 \pm 0.050$
      & $0.0094 \pm 0.0060$ \\
    \rowcolor{black!4}
    DriftWorld
      & $20.83 \pm 2.44$ & $0.731 \pm 0.072$ & $0.266 \pm 0.061$ & $0.0116 \pm 0.0064$ \\
    \rowcolor{cyan!6}
    \textbf{\method{} (Ours)}
      & $\bm{24.37 \pm 2.60}$
      & $\bm{0.839 \pm 0.054}$
      & $\bm{0.147 \pm 0.049}$
      & $\bm{0.0051 \pm 0.0036}$ \\
    \midrule
    \rowcolor{black!8}
    \multicolumn{5}{c}{\textit{3D Game Simulation: CSGO}} \\
    \midrule
    Model & PSNR $\uparrow$ & SSIM $\uparrow$ & LPIPS $\downarrow$ & MAE $\downarrow$ \\
    \midrule
    DIAMOND
      & $\underline{17.04 \pm 3.20}$
      & $\underline{0.513 \pm 0.108}$
      & $0.576 \pm 0.080$
      & $\underline{0.1187 \pm 0.0525}$ \\
    \rowcolor{black!4}
    Vid2World
      & $16.69 \pm 3.14$
      & $0.487 \pm 0.108$
      & $\underline{0.518 \pm 0.084}$
      & $0.1234 \pm 0.0509$ \\
    \rowcolor{cyan!6}
    \textbf{\method{} (Ours)}
      & $\bm{18.36 \pm 3.02}$
      & $\bm{0.565 \pm 0.097}$
      & $\bm{0.516 \pm 0.086}$
      & $\bm{0.0934 \pm 0.0358}$ \\
    \midrule
    \rowcolor{black!8}
    \multicolumn{5}{c}{\textit{Embodied Navigation: RECON}} \\
    \midrule
    Model & PSNR $\uparrow$ & SSIM $\uparrow$ & LPIPS $\downarrow$ & MAE $\downarrow$ \\
    \midrule
    NWM
      & $13.03 \pm 1.46$
      & $0.410 \pm 0.082$
      & $0.507 \pm 0.051$
      & $0.1623 \pm 0.0359$ \\
    \rowcolor{black!4}
    RAE-NWM
      & $14.66 \pm 1.56$
      & $0.424 \pm 0.080$
      & $0.458 \pm 0.046$
      & $0.1285 \pm 0.0280$ \\
    Vid2World
      & $\underline{15.59 \pm 1.74}$
      & $\underline{0.488 \pm 0.080}$
      & $\underline{0.416 \pm 0.054}$
      & $\underline{0.1099 \pm 0.0262}$ \\
    \rowcolor{cyan!6}
    \textbf{\method{} (Ours)}
      & $\bm{16.70 \pm 1.67}$
      & $\bm{0.546 \pm 0.076}$
      & $\bm{0.401 \pm 0.050}$
      & $\bm{0.0895 \pm 0.0221}$ \\
    \bottomrule
  \end{tabular*}
  \vspace{-10pt}
\end{table}

\begin{figure}[t]
  \centering
  \includegraphics[width=\linewidth]{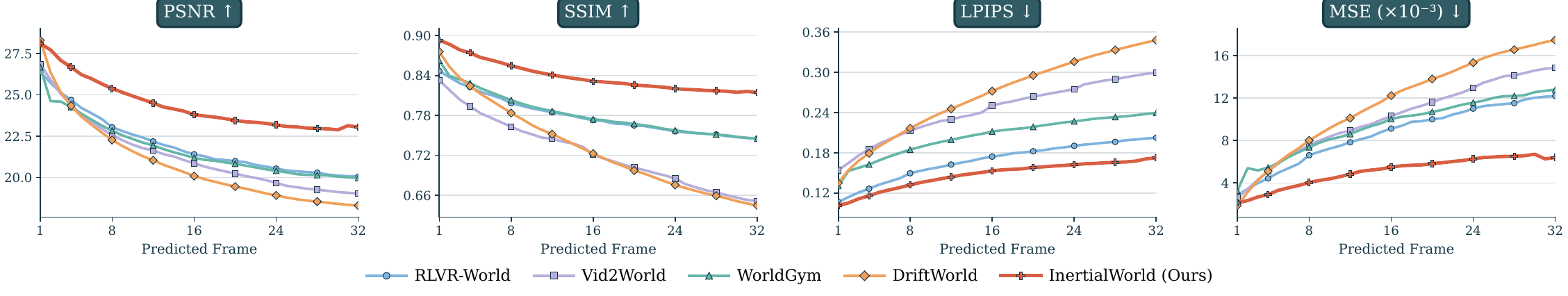}
  \vspace{-15pt}
  \caption{\textbf{Prediction quality throughout the 32-step RT-1 rollout.} Each curve reports the episode-level mean at every predicted frame under the same 1,024-episode protocol as Table~\ref{tab:main_results}.}
  \vspace{-18pt}
  \label{fig:rt1_metrics_over_time}
\end{figure}

\begin{figure}[t]
  \centering
  \includegraphics[width=\linewidth]{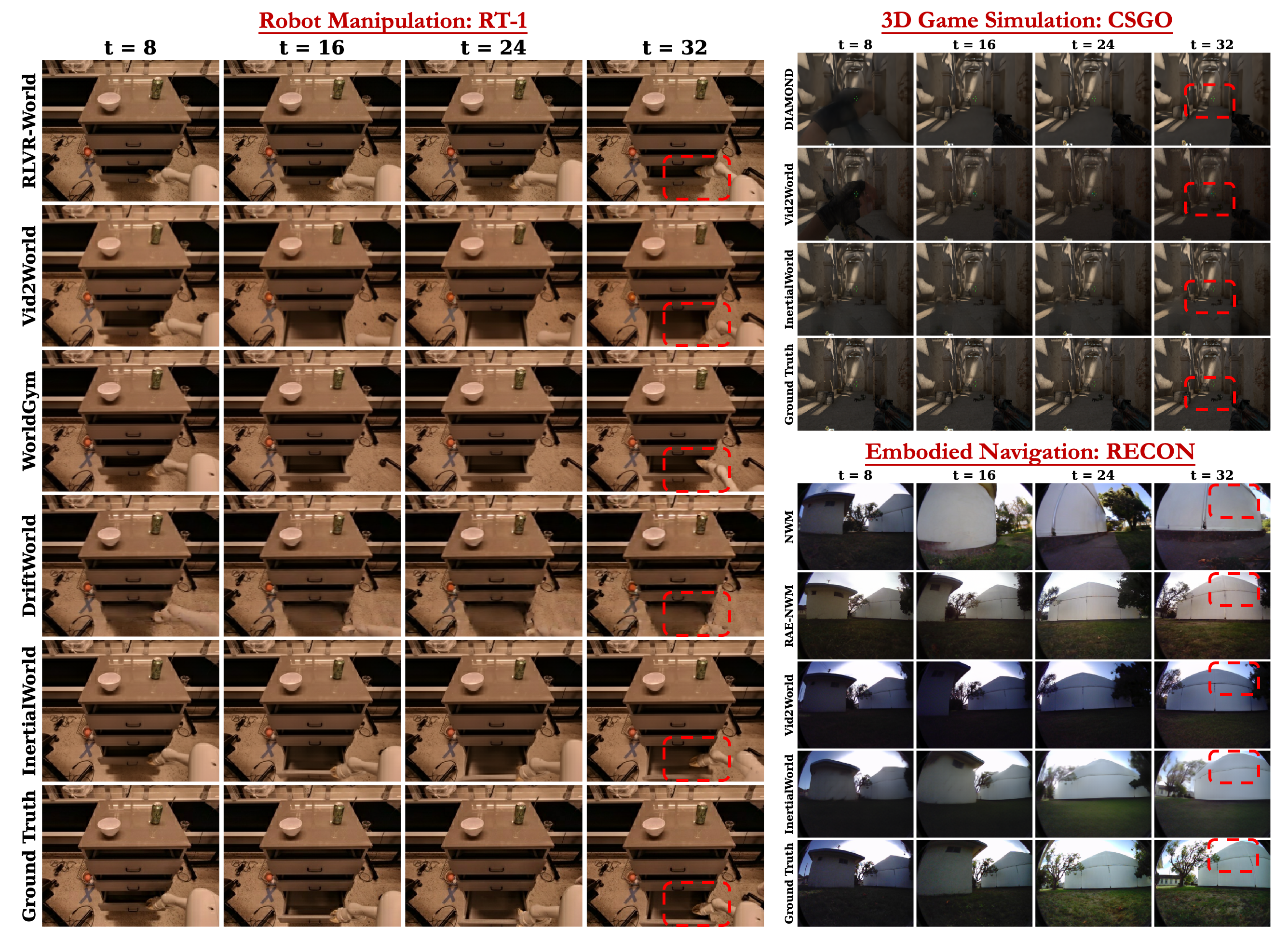}
  \vspace{-20pt}
  \caption{\textbf{Qualitative comparison across the three evaluation domains.} We show predicted frames at steps 8, 16, 24, and 32 for RT-1 robot manipulation, CSGO game simulation, and RECON embodied navigation. Each method uses its native conditioning interface, while all rows within a domain share the same episode, anchor, and action sequence. Red boxes mark structurally informative regions at the final step.}
  \label{fig:qualitative_three_domains}
  \vspace{-15pt}
\end{figure}

\noindent\textbf{Temporal Stability.}
Figure~\ref{fig:rt1_metrics_over_time} resolves the aggregate RT-1 results over the complete autoregressive rollout. \method{} achieves the highest PSNR and SSIM and the lowest LPIPS and MSE at every predicted frame. Its advantage is retained as the prediction horizon increases, showing that the main-table gain reflects consistently more stable video evolution rather than an improvement confined to the first few transitions.

\noindent\textbf{Qualitative Results.}
Figure~\ref{fig:qualitative_three_domains} provides a complementary visual comparison under the same 32-step protocols. On RT-1, \method{} preserves the articulated drawer geometry and the spatial relation between the drawer and robot through the final prediction. On CSGO, it retains the corridor layout and action-induced scene evolution without the pronounced structural degradation visible in competing rollouts. On RECON, it better maintains the curved building boundary, ground plane, and surrounding vegetation as the camera advances. Across all three domains, the benefit is most apparent at later predicted frames and in the highlighted regions, where independently reconstructing persistent content tends to blur, duplicate, or displace scene structure.

\noindent\textbf{Quantitative Results.}
Table~\ref{tab:main_results} summarizes the released-checkpoint comparison. On RT-1, \method{} obtains the strongest values across pixel fidelity, structural similarity, and perceptual distance. Against the strongest baseline for each metric, PSNR increases from $21.95$ to $24.37$ dB, SSIM from $0.782$ to $0.839$, LPIPS decreases from $0.168$ to $0.147$, and MSE from $0.0086$ to $0.0051$. On the 70-episode CSGO evaluation, \method{} remains competitive across all four metrics. Against DIAMOND, PSNR increases from $17.04$ to $18.36$ dB, SSIM from $0.513$ to $0.565$, and MAE decreases from $0.1187$ to $0.0934$; against Vid2World, LPIPS decreases from $0.518$ to $0.516$. On the held-out RECON protocol, \method{} is likewise competitive across all reported metrics. Against Vid2World, PSNR increases from $15.59$ to $16.70$ dB, SSIM from $0.488$ to $0.546$, LPIPS decreases from $0.416$ to $0.401$, and MAE from $0.1099$ to $0.0895$. Across the three domains, all 36 baseline--metric point estimates favor \method{}. Episode-paired 95\% bootstrap intervals and paired sign-flip tests with joint Holm correction support 35 of these 36 comparisons, including 11 of the 12 strongest-baseline margins; the sole unresolved case is the small CSGO LPIPS difference against Vid2World (Appendix~\ref{app:main_table_paired}). RT-1 is our primary domain and is trained for 60k steps; because training separate domain models is costly, the CSGO and RECON models use a fixed 30k-step budget and serve as secondary tests of cross-domain applicability rather than exhaustively tuned domain-specific systems.
\paragraph{Long-Horizon Persistence.}
\begin{figure}[t]
  \centering
  \includegraphics[width=\linewidth]{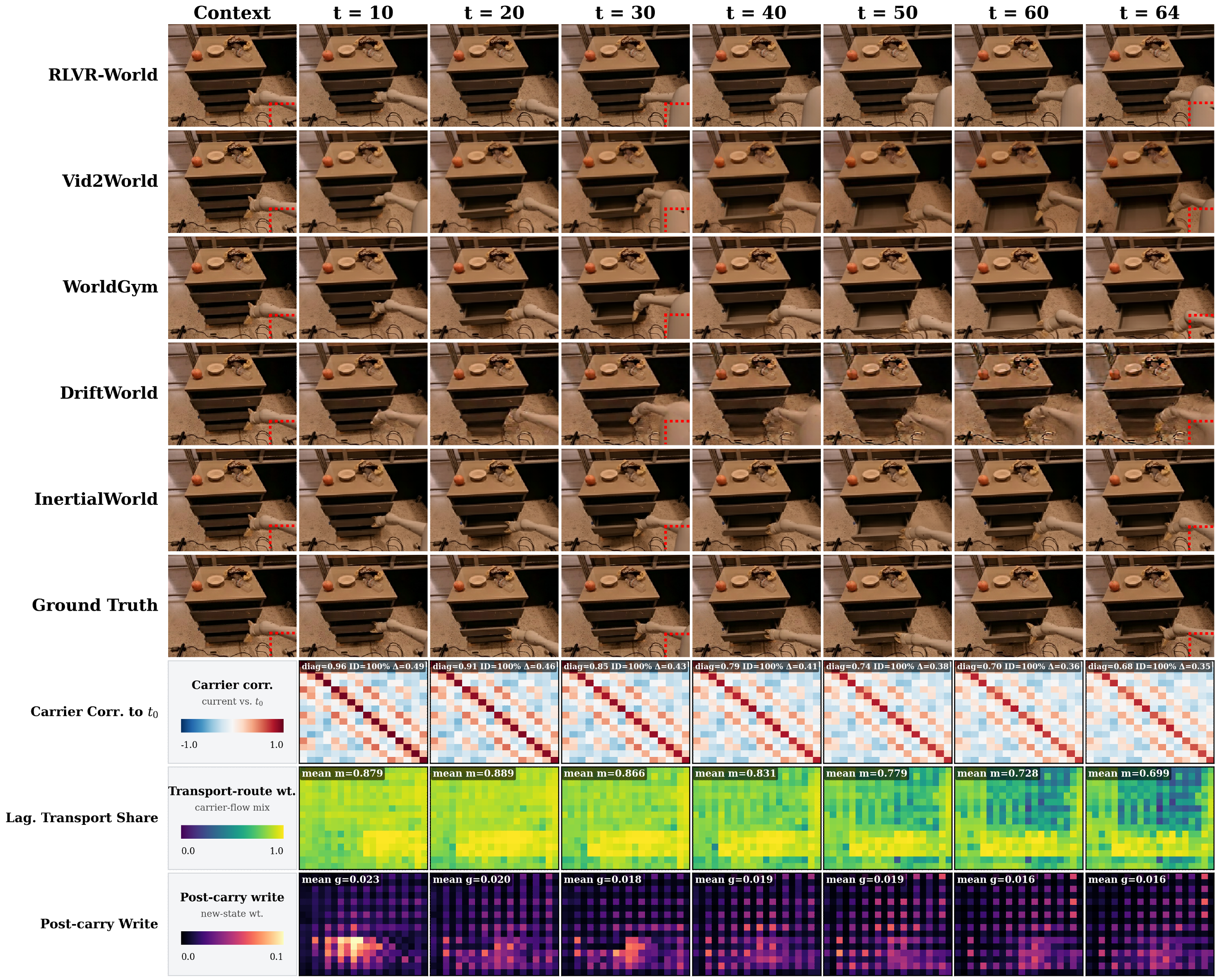}
  \vspace{-15pt}
  \caption{\textbf{Representative 64-step RT-1 case study.} Red boxes trace a local region from Context through occlusion around $t{=}30$ and its subsequent recovery, which remains preserved at $t{=}64$. Bottom rows show carrier correspondence, Lagrangian transport share, and post-carry write for \method{}.}
  \label{fig:carrier_transport_write}
  \vspace{-15pt}
\end{figure}

To probe beyond the maximum training horizon $H{=}32$, we roll out $H{=}64$ future frames, indexed by $t\in\{1,\ldots,H\}$ after Context. Exactly 35 of the 1,024 frozen test episodes contain 64 aligned future frames and actions; we evaluate all 35 without subsampling, with eligibility determined only by sequence length, and report the aggregate results in Appendix~\ref{app:rt1_h64}. Figure~\ref{fig:carrier_transport_write} is a representative mechanism case rather than the basis of this aggregate. A Context-visible structure is occluded by the robot arm around $t{=}30$, then reappears and remains preserved through $t{=}64$; \method{} most closely retains the recovered region, drawer geometry, and task progression, whereas the competing long-horizon rollouts deviate more visibly from the reference.

The bottom rows diagnose this state evolution. For the 16 carriers, the $16\!\times\!16$ cosine matrix compares every current carrier with every Context-initialized carrier. ``diag'' averages corresponding-slot similarities, ``ID'' is the fraction whose closest Context match has the same slot, and $\Delta$ averages the gap between that match and the strongest other-slot match; exact definitions are given in Eq.~\plaineqref{eq:app_carrier_correspondence}. From $t{=}10$ to $t{=}64$, diag decreases from $0.96$ to $0.68$ as content evolves, but ID remains $100\%$ and $\Delta$ remains positive ($0.49$ to $0.35$), showing adaptation without identity swaps. This carrier-space matrix is not an image heat map: its warm diagonal denotes same-slot correspondence. The other two rows are image-aligned $16\!\times\!20$ maps with shared scales over time. Transport share $m$ forms the committed flow $\mathbf{f}=(1-m)\mathbf{f}^{E}+m\mathbf{f}^{L}$, where $\mathbf{f}^{E}$ and $\mathbf{f}^{L}$ are Eulerian and carrier-derived Lagrangian flows; brighter cells indicate greater Lagrangian reliance, not greater motion magnitude. Its mean $\overline{m}$ remains high ($0.879$ to $0.699$), so carrier transport remains dominant but not exclusive. The post-carry write gate $g$ is the fraction of the new candidate committed after transport: $g{=}0$ retains the carried state, whereas $g{=}1$ replaces it with the candidate. Its spatial mean over all 320 cells is low ($\overline{g}=0.023$ to $0.016$) because most of the scene remains static, while localized bright responses around the robot--drawer interaction and newly exposed region show that the gate activates where new evidence must be incorporated. The transport map varies over the same interaction region, whereas the static background remains largely dark; together, the maps connect carrier transport and selective local writing to the visible motion and recovery above.

\subsection{Motion Fidelity under Action}
\label{sec:motion_fidelity}
\vspace{-3pt}

\noindent\textbf{Motion-Focused Diagnostics.}
Aggregate appearance metrics can be dominated by static regions, so we add three fixed diagnostics targeted at action-induced change. Because RT-1 lacks optical-flow annotations, the same frozen RAFT~\citep{teed2020raft} estimator measures flow in every prediction and reference rollout. \emph{Flow EPE} averages their endpoint discrepancy over all pixels and 32 transitions; it is not an error against measured flow ground truth. \emph{Flow Cosine} measures directional agreement on reference-dynamic pixels with flow magnitude of at least one pixel. \emph{Dynamic-region LPIPS} applies spatial LPIPS~\citep{zhang2018unreasonable} only within the same reference-motion mask. Each quantity is first aggregated within an episode and then across episodes.

\begin{figure}[t]
  \centering
  \includegraphics[width=\linewidth]{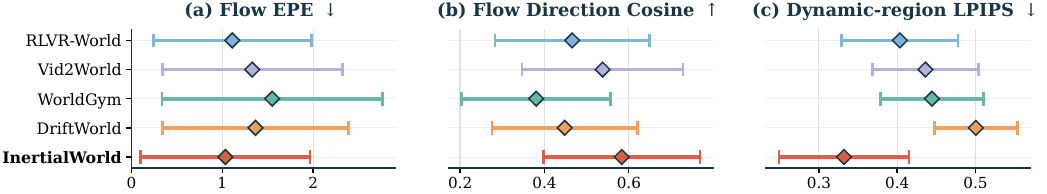}
  \vspace{-15pt}
  \caption{\textbf{GT-grounded motion fidelity on RT-1.} Colored diamonds denote episode means and capped horizontal bars denote one episode standard deviation over the same 1,024 episodes and 32 predicted frames as the RT-1 block of Table~\ref{tab:main_results}.}
  \label{fig:rt1_motion}
  \vspace{-15pt}
\end{figure}

\begin{table}[t]
  \centering
  \caption{\textbf{Paired motion-fidelity margins on RT-1.} \method{}-favorable paired differences and 95\% bootstrap intervals over 1,024 episodes. All 12 intervals exclude zero and survive Holm-corrected paired sign-flip tests ($p_{\mathrm{Holm}}=1.2{\times}10^{-4}$).}
  \label{tab:rt1_motion_paired}
  \vspace{0.3em}
  \small
  \renewcommand{\arraystretch}{1.13}
  {\color{black}
  \begin{tabular*}{\textwidth}{@{\extracolsep{\fill}}lccc@{}}
    \toprule
    Method
      & \shortstack{Flow EPE margin\\(95\% CI)}
      & \shortstack{Flow Cosine margin\\(95\% CI)}
      & \shortstack{Dyn. LPIPS margin\\(95\% CI)} \\
    \midrule
    RLVR-World
      & $0.076\ [0.054,\,0.097]$
      & $0.118\ [0.108,\,0.127]$
      & $0.071\ [0.067,\,0.075]$ \\
    \rowcolor{black!4}
    Vid2World
      & $0.296\ [0.264,\,0.327]$
      & $0.045\ [0.035,\,0.055]$
      & $0.104\ [0.100,\,0.108]$ \\
    WorldGym
      & $0.515\ [0.479,\,0.553]$
      & $0.203\ [0.193,\,0.213]$
      & $0.112\ [0.108,\,0.116]$ \\
    \rowcolor{black!4}
    DriftWorld
      & $0.332\ [0.304,\,0.360]$
      & $0.135\ [0.126,\,0.144]$
      & $0.169\ [0.165,\,0.172]$ \\
    \rowcolor{cyan!6}
    \textbf{\method{} (Ours)}
      & \multicolumn{3}{c}{\textbf{Reference method}} \\
    \bottomrule
  \end{tabular*}}
  \vspace{-20pt}
\end{table}

\noindent\textbf{Paired Motion Results.}
Figure~\ref{fig:rt1_motion} shows that \method{} ranks first on all three diagnostics, achieving the lowest Flow EPE and dynamic-region LPIPS together with the highest Flow Cosine. Table~\ref{tab:rt1_motion_paired} evaluates every baseline on the same episodes, rather than only the strongest baseline for each metric. Its margins are defined so that positive values always favor \method{}: baseline minus ours for the lower-is-better Flow EPE and dynamic-region LPIPS, and ours minus baseline for the higher-is-better Flow Cosine; the final blue row marks \method{} as the reference. All 12 paired margins have 95\% bootstrap intervals strictly above zero and remain significant after joint Holm correction, so every method--metric comparison favors \method{} under the paired analysis. The gains therefore persist when evaluation is concentrated on where and how the scene changes, rather than on the static pixels that dominate the frame. This behavior is consistent with the intended role of Inertial Memory: preserving stored frame content while action-conditioned motion transports it through the predicted video state.

\subsection{Real2Sim Policy Evaluation}
\label{sec:real2sim_policy}
\vspace{-3pt}

\begingroup

\setlength{\intextsep}{2pt}
\setlength{\columnsep}{8pt}
\setlength{\abovecaptionskip}{2pt}
\setlength{\belowcaptionskip}{0pt}
\begin{wrapfigure}{r}{0.58\textwidth}
  \centering
  \includegraphics[width=\linewidth]{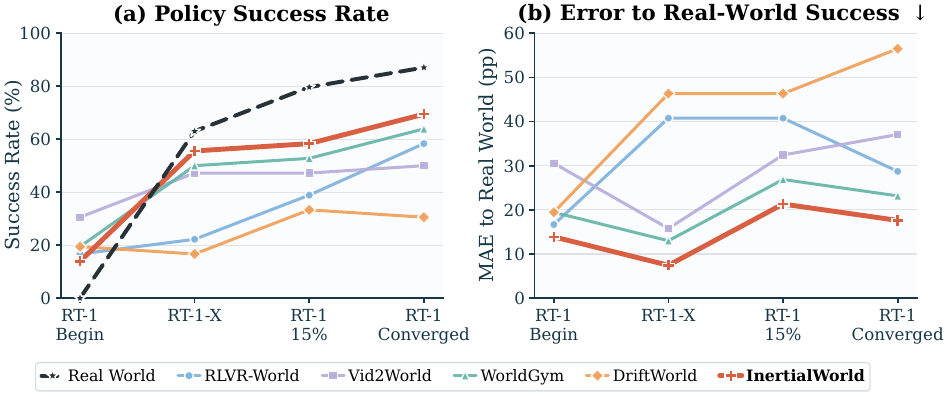}
  \caption{\textbf{Real2Sim policy evaluation.}}
  \label{fig:real2sim_policy}
\end{wrapfigure}

We test whether closed-loop predictions preserve differences in policy competence rather than only frame-level accuracy. The protocol crosses four frozen RT-1 policies (Begin, RT-1-X, RT-1 15\%, and Converged) with open- and close-drawer tasks and 18 requests per policy--task cell, yielding 144 trajectories per method and 720 across the five world models. All rollouts start from the same current observation; thereafter, the policy receives only the frame generated by that world model, and its predicted action conditions the next transition for at most 113 control steps. Each trajectory is scored under the same final-state task-success criterion; complete initialization, rollout, and annotation details appear in Appendix~\ref{app:real2sim_protocol}. Figure~\ref{fig:real2sim_policy}(a) reports success pooled equally over the two tasks for each policy, while panel (b) reports the task-averaged absolute gap to the SIMPLER real-world reference~\citep{li2024evaluating}, so opposite task errors cannot cancel. \method{} has the smallest gap for every policy and a $15.1$-percentage-point overall error, compared with $20.6$ for WorldGym, $29.0$ for Vid2World, $31.7$ for RLVR-World, and $42.1$ for DriftWorld. Its pooled success rises from $13.9\%$ for Begin to $55.6\%$, $58.3\%$, and $69.4\%$ for the stronger policies, preserving their competence ordering rather than uniformly inflating success. Because the references are external physical-rollout aggregates, we interpret these gaps as policy-level calibration rather than paired trajectory agreement. \par\vspace{-5pt} \endgroup

\begin{table}[!t]
  \centering
  \color{black}
  \vspace{-5pt}
  \caption{\textbf{Matched persistence ablations at 30k steps.} Episode mean $\pm$ sample standard deviation; lower is better. All variants are trained independently from scratch.}
  \label{tab:rt1_ablation_30k}
  \normalsize
  \setlength{\tabcolsep}{0.5pt}
  \renewcommand{\arraystretch}{1.10}
  \begin{tabular*}{\linewidth}{@{}l@{\extracolsep{\fill}}ccc@{}}
    \toprule
    Model
      & \shortstack{Long-lived\\content $\downarrow$}
      & \shortstack{Post-reappearance\\content $\downarrow$}
      & \shortstack{Late-track\\content $\downarrow$} \\
    \midrule
    \rowcolor{black!8}
    \multicolumn{4}{c}{\textit{RT-1, seed 0}} \\
    \mbox{Plain AR}
      & $0.5144 \pm 0.1275$ & $0.5144 \pm 0.1538$
      & $0.1009 \pm 0.0558$ \\
    \rowcolor{black!4}
    \mbox{Recurrent Memory}
      & $0.4414 \pm 0.1289$ & $0.4375 \pm 0.1525$
      & $0.0935 \pm 0.0565$ \\
    \mbox{Dense-only}
      & $0.4247 \pm 0.1334$ & $0.4183 \pm 0.1637$
      & $0.0896 \pm 0.0573$ \\
    \rowcolor{cyan!6}
    \mbox{\textbf{\method{} (Full)}}
      & $\bm{0.4145 \pm 0.1328}$ & $\bm{0.4036 \pm 0.1620}$
      & $\bm{0.0876 \pm 0.0552}$ \\
    \midrule
    \rowcolor{black!8}
    \multicolumn{4}{c}{\textit{RT-1, seed 1}} \\
    \mbox{Plain AR}
      & $0.5172 \pm 0.1272$ & $0.5073 \pm 0.1569$
      & $0.1051 \pm 0.0579$ \\
    \rowcolor{cyan!6}
    \mbox{\textbf{\method{} (Full)}}
      & $\bm{0.4153 \pm 0.1354}$ & $\bm{0.4053 \pm 0.1644}$
      & $\bm{0.0888 \pm 0.0561}$ \\
    \midrule
    \rowcolor{black!8}
    \multicolumn{4}{c}{\textit{CSGO, seed 0}} \\
    \mbox{Plain AR}
      & $0.6076 \pm 0.0761$ & $0.5961 \pm 0.0915$
      & $0.4864 \pm 0.1211$ \\
    \rowcolor{cyan!6}
    \mbox{\textbf{\method{} (Full)}}
      & $\bm{0.5679 \pm 0.0917}$ & $\bm{0.5624 \pm 0.1076}$
      & $\bm{0.4485 \pm 0.1200}$ \\
    \bottomrule
  \end{tabular*}
  \vspace{-19pt}
\end{table}

\subsection{Ablation Study}
\label{sec:memory_ablation}
% \vspace{-2pt}

We train four 0.8B-scale models from scratch for 30k updates under a matched protocol. The Full model combines dense Eulerian state with persistent Lagrangian carriers; \emph{Dense-only} removes the carriers, canonical content, and transport while retaining the Eulerian state; \emph{Recurrent Memory} replaces this structured state with a conventional GRU; and \emph{Plain AR} uses no recurrent memory. All variants share the tokenizer, data, action representation, optimizer, AR8$\rightarrow$AR16$\rightarrow$AR32 curriculum, global batch size, and ordered 1,024-episode H32 protocol. We evaluate content persistence on frozen DINO tracks defined solely from ground truth: long-lived error measures displaced Context content over the fixed late half of the rollout (Steps 17--32), post-reappearance error evaluates content after a ground-truth gap of at least four steps anywhere in H32, and late-track error covers all reliable tracks in the same late-half window. The main evaluation covers all 32 steps; this shared midpoint window is used only to emphasize accumulated state persistence rather than short-range prediction quality. For all three prediction-to-reference feature errors, lower is better; they target persistence failures that can be obscured by static regions in full-frame metrics. On all three measures in Table~\ref{tab:rt1_ablation_30k}, the Full model ranks first, followed by Dense-only, Recurrent Memory, and Plain AR. The Full model also improves over Dense-only in every case; all three pointwise paired 95\% percentile-bootstrap intervals exclude zero and are reported in Appendix Table~\ref{tab:app_ablation_paired}. The larger improvements over Recurrent Memory and Plain AR further favor structured inertial state over generic recurrence or direct autoregression, and the Full--Plain advantage repeats under an independent RT-1 seed and on CSGO. Together with Figure~\ref{fig:carrier_transport_write}, these results demonstrate that our proposed Inertial Memory preserves evolving content more effectively than dense recurrence, generic recurrence, or direct autoregression. Appendix~\ref{app:ablation_details} gives the complete protocol and paired statistics.

%% file: sec/5_conclusions.tex
\section{Conclusion}
\label{sec:conclusion}

In this work, we introduce \method{}, an Inertial Video World Model that treats prediction as the evolution of an existing video state rather than its repeated reconstruction. Inertial Memory realizes this principle through two complementary views: a dense Eulerian grid preserves scene-aligned evidence, while persistent Lagrangian carriers bind and transport regional content under action-conditioned motion. The model then predicts the complete next state and uses learned gates to incorporate changes that transport alone cannot explain. This carry-before-write transition makes persistence the default while retaining the flexibility required for newly visible content and non-rigid change. Across paired 32-step evaluations on RT-1, CSGO, and RECON, \method{} achieves the strongest aggregate prediction results and improves diagnostics focused on action-induced motion. Controlled ablations and mechanism diagnostics further support the roles of structured transport and state-dependent memory updates. Together, these results identify inertial state evolution as a useful design principle for interactive video world models. Extending the current deterministic, learned-coordinate formulation to multimodal futures and explicit three-dimensional scene representations remains an important direction.

%% file: sec/X_suppl.tex
\section{Complete Formulation of Inertial Memory}
\label{app:inertial_memory}

This section expands the compact formulation in Section~\ref{sec:inertial_memory}. We separate the state that implements the proposed inertial prior from auxiliary predictor context, and then describe a complete causal transition. Batch dimensions are omitted. Every recurrent variable is a function only of the initial observation, the action prefix, and previous model predictions.

\subsection{Frozen Video Representation}
\label{app:frozen_representation}

The frozen video tokenizer decomposes a video into a rollout-level Context code $\mathbf{C}$ and frame-level Dynamic codes $\mathbf{D}_{\tau}$. The Context code contains two $32\!\times\!40$ lanes and is computed once from $\mathbf{o}_t$. A Dynamic code contains four $8\!\times\!10$ lanes that represent base and detail components of motion and appearance. It is a complete discrete representation of a frame, rather than a residual relative to an observed image. During a rollout, $\mathbf{C}$ remains fixed and every future $\widehat{\mathbf{D}}_{\tau}$ is predicted by the model.

Let $\tau$ index recurrent rollout steps. The recurrent video state is
\begin{equation}
  \mathcal{S}_{\tau}=(\mathbf{C},\mathcal{I}_{\tau},
  \mathbf{H}_{\tau},\mathbf{Q}_{\tau}),\qquad
  \mathcal{I}_{\tau}=(\mathbf{G}_{\tau},
  \{\mathcal{R}_{\tau,k}\}_{k=1}^{K},\mathbf{B}_{\tau}).
  \label{eq:app_complete_state}
\end{equation}
In Eq.~\plaineqref{eq:app_complete_state}, $\mathcal{I}_{\tau}$ is Inertial Memory and $\mathcal{R}_{\tau,k}$ is the complete recurrent state of carrier $k$. Its Eulerian component $\mathbf{G}_{\tau}$ is a dense $16\!\times\!20$ grid aligned with scene coordinates. Its Lagrangian component contains $K=16$ persistent regional carriers. Carrier $k$ maintains a descriptor, a canonical feature map and support, a translation--rotation--scale coordinate, persistent velocity, visibility and existence estimates, and a bounded local deformation. The soft binding map $\mathbf{B}_{\tau}$ assigns Eulerian evidence between the carriers and a dense background branch. The carriers are learned regions, not supervised semantic objects, and the background branch avoids imposing a complete object decomposition.

The implementation additionally keeps the four most recent \emph{predicted} Dynamic states $\mathbf{H}_{\tau}$ and 64 generic interaction slots $\mathbf{Q}_{\tau}$. They provide predictor context inherited from the video backbone; the proposed persistent video state and its carry operation are defined by $\mathcal{I}_{\tau}$. This distinction is important: the method is not equivalent to retaining a longer image history as passive attention context.

\subsection{Initialization and Dual-View Binding}
\label{app:initialization_binding}

Learned projections of $\mathbf{C}$ initialize the dense grid and interaction slots. The first $K$ slots initialize the carrier descriptors. Carrier coordinates begin on a normalized $4\!\times\!4$ lattice with zero velocity; their canonical content is sampled from the initialized dense state. This constructs $\mathcal{S}_0=\operatorname{Init}(\mathbf{C})$ from the single observed frame, without an object detector or calibrated geometry.

At each recurrent step, a carrier is rendered into scene coordinates using its canonical content, support, coordinate, rigidity, and deformation. Its binding score at an Eulerian cell combines content compatibility, rendered support, visibility, and existence. A softmax over the $K$ carriers and the background branch gives $\mathbf{B}_{\tau}$. The dense view consequently retains globally available scene evidence, whereas the carrier view gives local content a persistent coordinate system in which it can move.

\subsection{Action-Conditioned Carry}
\label{app:action_carry}

The current action is combined with every carrier state, and a learned contact graph propagates action-dependent interactions between nearby carriers. The dynamics head predicts bounded changes to carrier velocity, coordinate, rigidity, existence, visibility, and deformation. Abstractly, the persistent coordinate update is
\begin{equation}
  \mathbf{v}_{\tau+1,k}=\eta\mathbf{v}_{\tau,k}
  +\Delta\mathbf{v}_{\tau,k},\qquad
  \boldsymbol{\xi}_{\tau+1,k}=\Pi(
  \boldsymbol{\xi}_{\tau,k}+\mathbf{v}_{\tau+1,k}),
  \label{eq:app_carrier_dynamics}
\end{equation}
Here $\eta=0.90$ is the fixed velocity-retention coefficient, $\Delta\mathbf{v}_{\tau,k}$ is the bounded action- and contact-conditioned learned increment, and $\boldsymbol{\xi}_{\tau,k}$ concatenates 2-D translation, rotation, and log scale. In Eq.~\plaineqref{eq:app_carrier_dynamics}, $\Pi$ wraps rotation and projects translation and log scale to their valid ranges. Carrier motion is therefore accumulated through a persistent state rather than independently re-estimated from scratch.

Old and updated carrier coordinates induce a Lagrangian correspondence. At Eulerian cell $i$, the binding-weighted carrier flow $\mathbf{f}^{L}_{\tau,i}$ is combined with dense fallback flow $\mathbf{f}^{E}_{\tau,i}$ according to foreground binding mass $m_{\tau,i}=\sum_k B_{\tau,ik}$, where $B_{\tau,ik}$ is the soft binding of cell $i$ to carrier $k$:
\begin{equation}
  \mathbf{f}_{\tau,i}=(1-m_{\tau,i})\mathbf{f}^{E}_{\tau,i}
  +m_{\tau,i}\mathbf{f}^{L}_{\tau,i}.
  \label{eq:app_hybrid_flow}
\end{equation}
Eq.~\plaineqref{eq:app_hybrid_flow} defines $m_{\tau,i}\in[0,1]$ as a Lagrangian mixture share, not a motion magnitude. The resulting correspondence carries owned source features with normalized bilinear forward splatting. Source ownership and destination coverage are kept separate: moved content is vacated at its source, while a learned reveal proposal fills newly uncovered locations. This prevents a feature from being duplicated at both its old and new positions. Zero predicted motion reduces to identity carry. We denote the complete inherited proposal by $\widetilde{\mathcal{S}}_{\tau+1}$.

\subsection{Complete Prediction and Gated Commit}
\label{app:prediction_commit}

The predictor reads the fixed Context code, carried dense state, rendered carriers, predicted-state history, interaction slots, and current action. It predicts all four lanes of $\widehat{\mathbf{D}}_{\tau+1}$. For each lane and cell, the readout assigns probability to three sources: newly generated content, stationary content from the previous predicted state, and content transported from a previous predicted location. Invalid persistence routes are masked. At evaluation time, the hard route is sampled deterministically from the learned posterior using an immutable key derived from the Context code, recurrent step, lane, and cell. No route has access to a future target.

Prediction and recurrent storage are deliberately distinct. Let $\mathcal{M}$ collect the writable fields of Inertial Memory and any predictor state updated by the transition. In the compact notation of Section~\ref{sec:inertial_memory},
\begin{equation}
  (\widehat{\mathbf{D}}_{\tau+1},\mathcal{M}^{\mathrm{new}}_{\tau+1},
  \mathbf{g}_{\tau+1})=P_{\theta}(\widetilde{\mathcal{S}}_{\tau+1},
  \mathbf{a}_{\tau};\mathbf{C}),\qquad
  \mathcal{M}_{\tau+1}=(1-\mathbf{g}_{\tau+1})\odot
  \widetilde{\mathcal{M}}_{\tau+1}+\mathbf{g}_{\tau+1}\odot
  \mathcal{M}^{\mathrm{new}}_{\tau+1}.
  \label{eq:app_commit}
\end{equation}
In Eq.~\plaineqref{eq:app_commit}, $P_{\theta}$ is the predictor with parameters $\theta$, $\mathbf{a}_{\tau}$ is the current action, $\widetilde{\mathcal{M}}_{\tau+1}$ denotes writable memory after carrying, $\mathcal{M}^{\mathrm{new}}_{\tau+1}$ is its newly predicted candidate, $\mathbf{g}_{\tau+1}\in[0,1]$ is the element-wise write gate, and $\odot$ denotes element-wise multiplication. A gate value of zero retains the carried state and a value of one replaces it with the candidate. The predicted Dynamic code reconstructs the complete next frame. Carrier descriptors are updated from their assigned dense evidence, and canonical content is updated by sampling the new dense state back into each carrier's coordinate system. The binding map, interaction slots, and predicted-state history are updated last.

One transition therefore follows the fixed order \emph{advance carriers $\rightarrow$ carry existing content $\rightarrow$ predict the complete next video state $\rightarrow$ commit memory}. This is the operational meaning of carry-before-write. Training targets define losses but never replace any field of $\mathcal{S}_{\tau}$. Hard Dynamic IDs are used on the forward path, straight-through surrogates provide gradients through discrete decisions, and the recurrence is not detached between steps. Training and evaluation thus execute the same predicted-only state evolution.

\section{Architecture and Training Details}
\label{app:architecture_training}

\subsection{Frozen Video Tokenizer}
\label{app:tokenizer}

We train a separate tokenizer for each domain while keeping its learnable architecture, objective, and optimization recipe identical. The copies differ only in their data readers and image preprocessing. Each tokenizer contains 120.386M parameters, uses 384 hidden and 256 latent channels, and maps a $256\!\times\!320$ internal canvas to the shared interface in Table~\ref{tab:app_architecture}. Every token is represented by five finite-scalar-quantization digits with levels $(7,5,5,5,5)$, giving 4,375 joint states~\citep{mentzer2024finite}.

\begin{table}[!htbp]
  \centering
  \caption{\textbf{Shared architecture of the frozen tokenizer and video world model.} Parameter counts exclude the frozen tokenizer.}
  \label{tab:app_architecture}
  \vspace{0.5em}
  \normalsize
  \setlength{\tabcolsep}{2pt}
  \renewcommand{\arraystretch}{1.12}
  \begin{tabular*}{\linewidth}{@{\extracolsep{\fill}}>{\raggedright\arraybackslash}m{0.25\linewidth}>{\raggedright\arraybackslash}m{0.68\linewidth}@{}}
    \toprule
    Component & Configuration \\
    \midrule
    \rowcolor{black!8}
    \multicolumn{2}{c}{\textit{Frozen tokenizer interface}} \\
    Context code
      & Two $32\!\times\!40$ lanes; 2,560 tokens; fixed within a rollout. \\
    \rowcolor{black!4}
    Dynamic code
      & Four $8\!\times\!10$ lanes; 320 tokens per predicted frame. \\
    Token quantization
      & Five FSQ digits with levels $(7,5,5,5,5)$. \\
    \midrule
    \rowcolor{black!8}
    \multicolumn{2}{c}{\textit{World-model predictor}} \\
    Predictor backbone
      & Width 1,280; 20 attention heads; 15 transition layers; three
        appearance-refinement layers; two Context blocks; one interaction
        layer; MLP ratio 4; zero dropout. \\
    \rowcolor{black!4}
    Recurrent features
      & 320 dense scene tokens, 16 carriers of width 256, 64 interaction
        slots, and four recent predicted Dynamic states. \\
    \bottomrule
  \end{tabular*}
\end{table}

Tokenizer training uses 33-frame clips: one Context frame and 32 future frames. An update supervises the Context frame and eight deterministically distributed future frames, always including the first and last targets and covering all four temporal quarters; validation decodes all 32 targets. The objective combines pixel reconstruction, edge and high-frequency fidelity, two-level wavelet detail, SSIM, LPIPS-VGG~\citep{zhang2018unreasonable}, quantization balance, and lane-specialization losses. Motion-region and late-frame terms are introduced at fixed stages after the initial reconstruction phase. We use no adversarial loss. This objective is identical across the three domains.

All tokenizers are trained from scratch for 60k optimizer steps with AdamW, learning rate $10^{-4}$, weight decay $0.02$, and $(\beta_1,\beta_2)=(0.9,0.95)$. The learning rate is warmed up for 2k steps and cosine-decayed to zero. Training uses BF16, global batch size 32, and gradient-norm clipping at 1.0. The completed 60k tokenizer is then frozen for video-world-model training.

\subsection{World-Model Architecture and Objective}
\label{app:wm_architecture}

The three domain models share all architectural and loss settings except the action projection. RT-1 and CSGO consume their raw 13-D and 51-D actions. RECON consumes normalized local-frame displacement in $x$ and $y$ together with wrapped relative yaw. These input dimensions account for the small parameter-count differences in Table~\ref{tab:app_domain_training}.

The primary objective supervises all digits of the four complete Dynamic lanes and their joint code. It additionally supervises the innovation, stationary, and transported routes. A frozen-decoder objective on two frames per update supplies RGB, edge, frequency, chroma, motion-region, and temporal change signals. Memory-consistency terms supervise binding, transport, background preservation, carrier local-deformation magnitude, and contact consistency using detached frozen-tokenizer geometry together with model-internal consistency signals. The frozen geometry is used only to compute losses and is never inserted into the recurrent state. Transport, stationary routing, contact, and local deformation are staged in during the first 12k steps so that an untrained persistence path cannot dominate the initial predictor. The complete set of coefficients is identical across domains.

\noindent\textbf{Loss specification.}
We set $\lambda_{\mathrm{mem}}=\lambda_{\mathrm{reg}}=1$ in Eq.~\plaineqref{eq:training_objective}. Table~\ref{tab:app_complete_loss} expands the grouped objective into every additive constituent; no additional loss terms are used. The first five rows constitute $\mathcal{L}_{\mathrm{pred}}$; three-way route CE and all Eulerian--Lagrangian rows constitute $\mathcal{L}_{\mathrm{mem}}$; and slot diversity plus dense-flow regularization constitute $\mathcal{L}_{\mathrm{reg}}$. Because both group coefficients equal one, the row coefficients in the table are the final additive weights. Coefficients are shared across domains and, unless a ramp is listed, are active from the first optimizer step.

\begin{table*}[t!]
  \centering
  \color{black}
  \caption{\textbf{Complete world-model optimization recipe.} Each row gives the exact Full-model definition and coefficient, its staging, and whether the matched Plain AR control retains it. ``Shared'' denotes the same target, coefficient, and schedule.}
  \label{tab:app_complete_loss}
  \vspace{0.5em}
  \normalsize
  \setlength{\tabcolsep}{1.5pt}
  \renewcommand{\arraystretch}{0.98}
  \begin{tabular*}{\textwidth}{@{\extracolsep{\fill}}>{\raggedright\arraybackslash}m{0.16\textwidth}>{\raggedright\arraybackslash}m{0.40\textwidth}>{\raggedright\arraybackslash}m{0.23\textwidth}>{\raggedright\arraybackslash}m{0.11\textwidth}@{}}
    \toprule
    Objective & Exact definition and coefficient & Staging & Plain AR \\
    \midrule
    \rowcolor{black!8}
    \multicolumn{4}{c}{\textit{Token and visual prediction}} \\
    Joint-code CE
      & $0.50$ on the complete Dynamic joint ID
      & Active from step 0 & Shared \\
    \rowcolor{black!4}
    Generation-code CE
      & $0.10\,[0.25\text{ all-cell}+0.75\text{ innovation-cell}]$
      & Active from step 0 & $0.10$, all cells \\
    Digit CE
      & $0.25$ on all FSQ digits
      & Detail-lane factor $r_{\mathrm{detail}}{:}\ 0.25\!\rightarrow\!1$
        over 0--4k & Shared \\
    \rowcolor{black!4}
    Ordered-digit error
      & $0.05$ on the ordered FSQ value
      & Same detail-lane factor $r_{\mathrm{detail}}$ & Shared \\
    Frozen-decoder objective
      & $1.0$ RGB $+\,0.30$ edge $+\,0.30$ high frequency $+\,0.50$ chroma
        $+\,1.0$ motion-region $+\,0.15$ temporal residual
      & Off through 1k, then $\rightarrow1$ by 6k & Shared \\
    \midrule
    \rowcolor{black!8}
    \multicolumn{4}{c}{\textit{Route and recurrent regularization}} \\
    Three-way route CE
      & $0.10$ over innovation, stationary, and transported routes
      & Transport $0.1\!\rightarrow\!1$ by 6k; stationary off through 1k,
        then $\rightarrow1$ by 4k & Not used \\
    \rowcolor{black!4}
    Slot diversity
      & $0.002$ times mean squared off-diagonal slot cosine similarity
      & Active from step 0 & Not used \\
    Dense-flow regularizer
      & $0.0005$ times mean component-wise squared flow
      & Active from step 0 & Not used \\
    \midrule
    \rowcolor{black!8}
    \multicolumn{4}{c}{\textit{Eulerian--Lagrangian supervision}} \\
    Region composite
      & $0.010r_{\mathrm{aux}}$ times [moving-cell foreground NLL
        $+$ moving-cell assignment entropy normalized by $\log 16$
        $+\,0.20$ static-background NLL
        $+\,0.10$ squared foreground-mass excess above the target budget]
      & $r_{\mathrm{aux}}{:}\ 0.1\!\rightarrow\!1$ by 4k & Not used \\
    \rowcolor{black!4}
    Assignment consistency
      & $0.050r_{\mathrm{aux}}$ times motion/visibility-weighted L1 between
        the current 16-carrier-plus-background assignment and the previous
        assignment sampled at the predicted source grid
      & Same region ramp & Not used \\
    Carrier local-deformation magnitude
      & $0.010r_{\mathrm{aux}}$ times the carrier-mass-weighted squared RMS
        local-deformation magnitude
      & Same region ramp & Not used \\
    \rowcolor{black!4}
    Contact consistency
      & $0.002r_{\mathrm{aux}}r_{\mathrm{contact}}$ times
        [pseudo-target BCE with positive-class weight $4$
        $+\,0.05$ off-diagonal mass
        $+$ squared diagonal $+\,0.50$ symmetry L1]
      & $r_{\mathrm{contact}}{=}0$ through 2k, then $\rightarrow1$ by 8k
      & Not used \\
    Context source-grid error
      & $0.050r_{\mathrm{aux}}$ times weighted coordinate L1 to the frozen
        Context-referenced source grid
      & Same region ramp & Not used \\
    \rowcolor{black!4}
    Previous source-grid error
      & $0.100r_{\mathrm{aux}}$ times weighted coordinate L1 to the frozen
        adjacent-frame source grid
      & Same region ramp & Not used \\
    \bottomrule
  \end{tabular*}
\end{table*}

\noindent\textbf{Target construction and normalization.}
The frozen tokenizer supplies detached source grids, visibility, and motion; none enters the recurrent state. Its absolute source grid maps each current token to Context, and composing consecutive absolute grids gives the adjacent-frame source grid. Visible motion above $0.02$ seeds the foreground target after a $3\!\times\!3$ spatial dilation, and the preceding foreground is then advected causally through the adjacent grid. The static-background target is the visible, sub-threshold complement of this persistent foreground. The foreground budget is the target foreground fraction plus a $0.10$ margin. Assignment consistency compares the current background-plus-16-carrier distribution with the previous distribution bilinearly sampled at the model-predicted adjacent source grid. Its per-cell error is the mean absolute difference over these 17 channels, weighted by the product of frozen adjacent-motion gate and visibility and normalized by the total valid weight. Context-grid and adjacent-grid errors compare the two predicted grids with their respective frozen targets in normalized $[-1,1]^2$ coordinates; each coordinate L1 is normalized jointly over its batch, horizon, and cell supervision mass. Their weights are visibility times $[0.05+0.95(1-\exp(-\delta/0.15))]$, where $\delta$ is displacement in token units.

For carrier $i$, let $\mathbf{D}_i(u,v)\in\mathbb{R}^{2}$ be its raw local deformation at lattice site $(u,v)\in\{1,\ldots,4\}^{2}$. The quantity named the carrier local-deformation magnitude in Table~\ref{tab:app_complete_loss} is
\begin{equation}
  r_i=\left(\frac{1}{16}\sum_{u,v=1}^{4}
  \lVert\mathbf{D}_i(u,v)\rVert_2^2+10^{-12}\right)^{1/2}.
  \label{eq:app_local_deformation}
\end{equation}
With $\omega_i=\sum_c A_{ci}/\max(\sum_{j,c}A_{cj},10^{-8})$, the loss averages $\sum_i\omega_i r_i^2$ over the batch. Thus Eq.~\plaineqref{eq:app_local_deformation} regularizes the magnitude of local non-rigid deformation rather than a semantic-object label or a separate rigidity prediction; $A_{ci}$ is the carrier assignment defined below.

Contact pseudo-targets combine detached model assignments with frozen-tokenizer motion. Let $A_{ci}$ be the assignment of Eulerian cell $c$ to carrier $i$, $\mathbf{u}_c$ its frozen adjacent-frame motion, and $\mathbf{x}_c$ its identity-grid center. For $i,j\in\{1,\ldots,16\}$, define
\begin{equation}
\begin{gathered}
M_i=\max\!\left(\sum_c A_{ci},10^{-6}\right),\qquad
a_i=1-\exp(-\lVert\mathbf{v}_i\rVert_2/0.15),\qquad y_{ii}=0,\\
\mathbf{v}_i=\frac{\sum_c A_{ci}\mathbf{u}_c}{M_i},\qquad
\mathbf{p}_i=\frac{\sum_c A_{ci}\mathbf{x}_c}{M_i},\\
y_{ij}=\sqrt{a_i a_j}\,
\exp\!\left(-\frac{\lVert\mathbf{v}_i-\mathbf{v}_j\rVert_2}{0.20}\right)
\exp\!\left(-\frac{\lVert\mathbf{p}_i-\mathbf{p}_j\rVert_2^2}
{2(0.45)^2}\right),\qquad i\ne j.
\end{gathered}
\label{eq:app_contact_target}
\end{equation}
Here $M_i$ is assignment mass, $\mathbf{v}_i$ and $\mathbf{p}_i$ are its assignment-weighted motion and center, $a_i$ is motion activity, and $\lVert\cdot\rVert_2$ is the Euclidean norm. The complete target $\{y_{ij}\}$ is detached. Eq.~\plaineqref{eq:app_contact_target} therefore requires both activity and compatible motion at nearby locations rather than treating proximity alone as contact; its off-diagonal BCE uses positive-class weight $4$. The BCE and sparsity means use the ordered off-diagonal pairs, the diagonal term averages the 16 diagonal entries, and the symmetry term averages the full $16\!\times\!16$ matrix. Every loss is reduced over its valid cells or carriers before the scalar coefficient in Table~\ref{tab:app_complete_loss} is applied.

\noindent\textbf{Weighting and staged initialization.}
Let $s$ denote the optimizer step and $H$ the current curriculum rollout horizon. The four Dynamic lanes are normalized and averaged with weights $(1.0,0.80r_{\mathrm{detail}},1.25,0.80r_{\mathrm{detail}})$ for motion-base, motion-detail, appearance-base, and appearance-detail. The 15\% most changed cells receive weight $3$, and the final $\max(2,\lceil0.25H\rceil)$ rollout steps receive weight $1.5$. Every per-step structural term is multiplied by the same horizon weight and divided by the sum of horizon weights; source-grid errors instead use their explicit weighted supervision mass as the denominator. The visual objective decodes two adjacent frames per update: using zero-based target indices, even optimizer steps use $(H-2,H-1)$, whereas odd steps rotate the start index as $\lfloor s/2\rfloor\bmod(H-1)$. Its motion term evaluates the 12\% most changed target pixels. All ramps are linear between the stated endpoints and clipped outside them. The complete ramp definitions, with clipping made explicit, are
\begin{equation}
\begin{aligned}
\operatorname{clip}(x)&=\min(1,\max(0,x)),
&\qquad r_{\mathrm{detail}}(s)&=0.25+0.75\min(s/4000,1),\\
r_{\mathrm{aux}}(s)&=0.1+0.9\min(s/4000,1),
& r_{\mathrm{visual}}(s)&=\operatorname{clip}((s-1000)/5000),\\
r_{\mathrm{stationary}}(s)&=\operatorname{clip}((s-1000)/3000),
& r_{\mathrm{transport}}(s)&=0.1+0.9\min(s/6000,1),\\
r_{\mathrm{contact}}(s)&=\operatorname{clip}((s-2000)/6000),
& r_{\mathrm{def}}(s)&=\operatorname{clip}((s-4000)/8000).
\end{aligned}
\label{eq:app_staging_ramps}
\end{equation}
Thus region memory, dense-scene writes, and canonical-content updates reach full strength at 4k; Lagrangian transport at 6k; contact at 8k; and local deformation at 12k. Eq.~\plaineqref{eq:app_staging_ramps} also specifies the detail, frozen-decoder, and stationary-route schedules in Table~\ref{tab:app_complete_loss}. The local-deformation schedule is a structural multiplier on the deformation update and its induced source geometry, not an additional loss; its magnitude is regularized by the carrier-local-deformation-magnitude row in Table~\ref{tab:app_complete_loss}. Concretely, the state update retains $0.95$ of the previous field and adds $0.05$ of a tanh-bounded proposal with maximum scale $0.04$, multiplied once by the region and deformation strengths. The model-side region strength and loss-side $r_{\mathrm{aux}}$ use the same numerical schedule but are separate multipliers. We zero-initialize the scene-memory gate weights. For a non-vacated location at unit confidence, its keep-logit bias and keep floor give
\begin{equation}
  \kappa_0=\sigma(2.2)=0.900,\qquad
  \widetilde{\kappa}_0=0.8+0.2\kappa_0=0.980,\qquad
  g_0=1-\widetilde{\kappa}_0=0.020,
  \label{eq:app_gate_initialization}
\end{equation}
where $\sigma$ is the logistic sigmoid, $\kappa_0$ and $\widetilde{\kappa}_0$ are the raw and effective keep coefficients, and $g_0$ is the new-content write coefficient in Eq.~\plaineqref{eq:app_commit}. Vacated source locations bypass the keep floor so that transported content can leave its old position. The carrier-descriptor keep gate also has zero weights and bias $3.0$, giving $\sigma(3.0)=0.953$. Full-only targets use static-grid weight $0.05$, foreground-budget margin $0.10$, motion-gate scale $0.15$, static-motion threshold $0.02$, and transport-visibility floor $0.10$; contact pseudo-targets use motion/co-motion/proximity scales $(0.15,0.20,0.45)$.

\noindent\textbf{Matched Plain AR control.}
Plain AR retains the complete token-prediction and decoded-visual recipe, including the lane, changed-cell, terminal-frame, and visual schedules. Because it has no route variable, its generation CE is evaluated on all cells. All rows marked ``Not used''---including route, slot/flow, carrier, transport, contact, and source-grid objectives---are absent, with no replacement auxiliary objective. The frozen representations used to construct Full-only targets supervise losses only and are never written into recurrent state.

\begin{table}[t!]
  \centering
  \caption{\textbf{Domain-specific preprocessing and video-world-model training.} ``Quick/full'' gives the number of validation clips evaluated every 200/1,000 steps.}
  \label{tab:app_domain_training}
  \vspace{0.5em}
  \normalsize
  \setlength{\tabcolsep}{1.5pt}
  \renewcommand{\arraystretch}{1.12}
  \begin{tabular*}{\linewidth}{@{\extracolsep{\fill}}lcccc@{}}
    \toprule
    Action & \shortstack{WM\\params} & \shortstack{WM\\steps}
      & \shortstack{Peak-LR\\hold} & \shortstack{Quick/\\full} \\
    \midrule
    \rowcolor{black!8}
    \multicolumn{5}{c}{\textit{Robot Manipulation: RT-1}} \\
    raw 13-D & 818.746M & 60k & 2k--32k & 64/1,024 \\
    \midrule
    \rowcolor{black!8}
    \multicolumn{5}{c}{\textit{3D Game Simulation: CSGO}} \\
    raw 51-D & 818.795M & 30k & 2k--16k & 64/128 \\
    \midrule
    \rowcolor{black!8}
    \multicolumn{5}{c}{\textit{Embodied Navigation: RECON}} \\
    local $\Delta$ 3-D & 818.733M & 30k & 2k--16k & 13/13 \\
    \bottomrule
  \end{tabular*}
\end{table}

All video world models use AdamW with learning rate $10^{-4}$, minimum learning rate $10^{-5}$, weight decay $0.05$, and $(\beta_1,\beta_2)=(0.9,0.95)$. After a 2k-step warmup, the rate is held at its peak for the interval in Table~\ref{tab:app_domain_training} and then cosine-decayed to the minimum at the final step. The predicted-only rollout curriculum is shared: AR8 for steps $[0,4\mathrm{k})$, AR16 for $[4\mathrm{k},12\mathrm{k})$, and AR32 thereafter. We backpropagate through the entire active rollout without teacher forcing or truncated BPTT.

Formal training uses seed 0, BF16, PyTorch distributed data parallel over 32 GPUs, global batch size 32, no gradient accumulation, activation checkpointing, and gradient clipping at 1.0. Quick validation is run every 200 steps, full AR32 validation and checkpointing every 1,000 steps. The main results use the final raw-model checkpoint---60k for RT-1 and 30k for CSGO and RECON---rather than choosing a checkpoint on the test set. No exponential moving average is used. The ablation in Section~\ref{sec:memory_ablation} instead fixes every compared model to 30k steps.

\section{Datasets and Frozen Evaluation Protocols}
\label{app:datasets_protocols}

All protocols were materialized as ordered episode lists before metric aggregation. A protocol records the source episode, anchor, action slice, conditioning frames, and 32 reference targets. Every method receives the same anchor and the same actions for the 32 scored transitions. A baseline may additionally use frames and aligned actions strictly before the anchor when required by its released interface, but no method observes any scored target. Table~\ref{tab:app_protocols} summarizes the resulting contracts.

\begin{table}[t!]
  \centering
  \caption{\textbf{Frozen evaluation protocols.} Context is excluded from all reported metrics.}
  \label{tab:app_protocols}
  \vspace{0.5em}
  \normalsize
  \setlength{\tabcolsep}{1pt}
  \renewcommand{\arraystretch}{1.12}
  \begin{tabular*}{\linewidth}{@{\extracolsep{\fill}}>{\centering\arraybackslash}m{0.19\linewidth}>{\centering\arraybackslash}m{0.12\linewidth}>{\centering\arraybackslash}m{0.20\linewidth}>{\raggedright\arraybackslash}m{0.38\linewidth}@{}}
    \toprule
    Evaluated episodes & Targets & Resolution & Common alignment \\
    \midrule
    \rowcolor{black!8}
    \multicolumn{4}{c}{\textit{Robot Manipulation: RT-1}} \\
    1,024 & 32 & $256\!\times\!320$
      & $\mathbf{o}_t,\mathbf{a}_{t:t+31}\mapsto
        \mathbf{o}_{t+1:t+32}$ \\
    \midrule
    \rowcolor{black!8}
    \multicolumn{4}{c}{\textit{3D Game Simulation: CSGO}} \\
    70 of 372 test & 32 & $256\!\times\!320$
      & Same current anchor and raw 51-D action sequence. \\
    \midrule
    \rowcolor{black!8}
    \multicolumn{4}{c}{\textit{Embodied Navigation: RECON}} \\
    128 & 32 & $240\!\times\!320$
      & Same current anchor and adjacent local-motion sequence. \\
    \bottomrule
  \end{tabular*}
\end{table}

\subsection{RT-1}
\label{app:rt1_protocol}

We construct a fixed 1,024-episode protocol from RT-1 \citep{brohan2022rt}. Episodes used by our development validation protocol are first excluded. Of the remaining episodes, 1,142 admit both a 16-frame observation prefix and 32 adjacent future targets. We sample 1,024 distinct episodes without replacement using seed 0 and then sort the selected episode identifiers. The anchor within an episode is generated deterministically from the episode identifier and constrained to retain the preceding 15 frames and all 32 future frames. The common current observation is $\mathbf{o}_t$; the raw 13-D action $\mathbf{a}_{t+j}$ is paired with target $\mathbf{o}_{t+j+1}$. Our model observes only $\mathbf{o}_t$, whereas a released baseline may use frames from the reserved prefix according to its native interface.

This protocol is disjoint from the data used to train and validate \method{}. The released baselines were trained with different RT-1 partitions, so it cannot simultaneously be a held-out split for every public checkpoint. In particular, 1,015 of the 1,024 episodes occur in the released RLVR-World training subset. We therefore use this block as a paired released-checkpoint comparison under a common rollout contract, not as a claim that every baseline is evaluated on its own held-out distribution.

\subsection{CSGO}
\label{app:csgo_protocol}

We follow the 5,500-episode CSGO corpus and the 500-episode held-out partition used by Vid2World~\citep{huang2025vid2world}. The first 128 held-out episodes are reserved for validation and the remaining 372 define the test pool. For each test episode, a deterministic per-episode key selects an anchor uniformly from indices 32 through 967. Thus, every episode retains at least 32 earlier frames for native baseline context and exactly 32 future targets. All models use the same anchor and raw 51-D action interval.

The CSGO block reported in Table~\ref{tab:main_results} is a fixed 70-episode slice of this 372-episode protocol: zero-based canonical request indices $0,5,\ldots,345$. We state this rule explicitly so that the subset can be reconstructed exactly.

The source frames are $150\!\times\!280$ BGR images. We convert them to RGB and apply one antialiased bilinear resize to $256\!\times\!320$. The same spatial transform is applied independently to every frame and does not alter the temporal pairing between $\mathbf{a}_{t}$ and $\mathbf{o}_{t+1}$.

\subsection{RECON}
\label{app:recon_protocol}

We use the official Vid2World RECON split over 11,835 trajectories \citep{shah2021rapid,huang2025vid2world}: 9,468 for training and 2,367 held out. We exclude the 17 held-out trajectories used for development validation, retain the 1,154 trajectories that admit the required native history and 32 adjacent targets, and deterministically select 128 distinct test episodes with seed 0. A second deterministic key selects one valid anchor per episode. Every method receives the same anchor and sequence of normalized local $x$--$y$ displacement and relative-yaw actions.

RECON images are decoded at $480\!\times\!640$, resized with antialiased bilinear interpolation to $240\!\times\!320$, and reflection-padded by eight pixels above and below to form the $256\!\times\!320$ internal training canvas. The padding is removed before serialization and scoring, so all reported RECON metrics use the undistorted $240\!\times\!320$ field of view. Predictions from baselines with a different native image size are transformed once to this common scoring resolution.

\subsection{Canonical Targets and Leakage Checks}
\label{app:canonical_targets}

For each domain, the reference frames are decoded once through the frozen data preprocessing contract and used as the canonical target tensor for every method. We verify ordered episode identifiers, anchor indices, action bytes, frame count, color order, resolution, and value range before computing a metric. For RECON, where independently installed image libraries can produce small interpolation differences, baseline predictions are paired with this single canonical target tensor after confirming that their episode and anchor metadata match. The predicted bytes are not modified by this pairing.

The scored tensor always contains the 32 future frames $\mathbf{o}_{t+1:t+32}$. Native conditioning frames, the common anchor, and any preceding history are excluded. Autoregressive feedback contains only model predictions. All methods share $\mathbf{a}_{t:t+31}$ for the scored transitions. Native multi-frame baselines may additionally consume actions aligned with their observed prefix, and blockwise baselines receive the aligned subsequence for the frames they jointly generate. No method receives a reference image from the scored interval or an action after the final predicted transition.

\subsection{Closed-Loop Real2Sim Policy Evaluation}
\label{app:real2sim_protocol}

\noindent\textbf{Protocol.}
We evaluate policy discrimination under one frozen closed-loop protocol for all five world models. The protocol contains four RT-1 checkpoints (Begin, RT-1-X, RT-1 15\%, and Converged), two drawer tasks (open and close), and 18 requests in every policy--task cell. This gives 144 trajectories per method and 720 generated trajectories overall. Every method may consume only the native visual context required by its released interface. The first policy decision is made from the same current observation. Thereafter, the policy observes only the frame generated by that world model, and its predicted action conditions the next transition. We provide neither future dataset images nor future dataset actions. A rollout executes at most 113 control transitions and stores 114 frames including the initial observation; if the policy terminates earlier, we retain its final post-action frame.

\noindent\textbf{Scoring and aggregation.}
Each generated trajectory is manually scored once under a common final-state criterion: the instructed drawer must be in the requested open or closed state in the retained final observation. Let $S_{m,p,k}$ denote the success rate for method $m$, policy $p$, and task $k\in\{\mathrm{open},\mathrm{close}\}$. Figure~\ref{fig:real2sim_policy} reports the balanced pooled rate $S_{m,p}=\frac{1}{2}\sum_k S_{m,p,k}$, equivalently the mean over the 36 trajectories for that policy. Its calibration panel reports $E_{m,p}=\frac{1}{2}\sum_k |S_{m,p,k}-S^{\mathrm{real}}_{p,k}|$ and $\overline{E}_m=\frac{1}{4}\sum_p E_{m,p}$, where $S^{\mathrm{real}}_{p,k}$ is the corresponding SIMPLER physical-rollout success-rate reference for policy $p$ and task $k$. Computing the absolute gap within each task before averaging prevents opposite open- and close-drawer errors from canceling.

\noindent\textbf{Results and interpretation.}
The pooled \method{} success rates are $13.9\%$, $55.6\%$, $58.3\%$, and $69.4\%$ from Begin through Converged, while the corresponding SIMPLER real-world references are $0.0\%$, $63.0\%$, $79.7\%$, and $87.1\%$ \citep{li2024evaluating}. \method{} has the smallest $E_{m,p}$ for every policy. Its overall task-averaged error is $15.1$ percentage points, compared with $20.6$ for WorldGym, $29.0$ for Vid2World, $31.7$ for RLVR-World, and $42.1$ for DriftWorld. The real-world values are external aggregates from separate physical trials rather than paired outcomes for these generated requests; accordingly, we interpret the comparison as policy-level calibration and discrimination, not trajectory-level sim-to-real agreement.

\section{Baseline Reproduction and Metric Definitions}
\label{app:baselines_metrics}

\subsection{Released-Checkpoint Rollout Contracts}
\label{app:baseline_rollouts}

We preserve each baseline's released observation interface, checkpoint, and autoregressive decomposition instead of forcing all methods into our single-frame state interface. Table~\ref{tab:app_baseline_rollouts} records the native context and how 32 future frames are produced. In every case, the last conditioning frame is the common anchor $\mathbf{o}_t$, and all feedback after initialization is predicted by that method.

\begin{table}[!htbp]
  \centering
  \caption{\textbf{Native baseline interfaces used in the paired 32-frame evaluation.} $Hk$ denotes $k$ conditioning frames available to a released model; it does not denote scored output.}
  \label{tab:app_baseline_rollouts}
  \vspace{0.5em}
  \normalsize
  \setlength{\tabcolsep}{2pt}
  \renewcommand{\arraystretch}{1.12}
  \begin{tabular*}{\linewidth}{@{\extracolsep{\fill}}>{\raggedright\arraybackslash}m{0.20\linewidth}>{\centering\arraybackslash}m{0.17\linewidth}>{\raggedright\arraybackslash}m{0.53\linewidth}@{}}
    \toprule
    Method & Native context & 32-frame rollout \\
    \midrule
    \rowcolor{black!8}
    \multicolumn{3}{c}{\textit{Robot Manipulation: RT-1}} \\
    RLVR-World & released H7 state
      & Four 7-frame recurrent blocks followed by a final 4-frame output
        chunk. \\
    \rowcolor{black!4}
    Vid2World & initial H1, then H9
      & One 15-frame generation, followed by 17 single-frame sliding-window
        generations. \\
    WorldGym & H1
      & Official one-step latent transition applied 32 times. \\
    \rowcolor{black!4}
    DriftWorld & H4
      & Official four-frame sliding window applied autoregressively for 32
        outputs. \\
    \midrule
    \rowcolor{black!8}
    \multicolumn{3}{c}{\textit{3D Game Simulation: CSGO}} \\
    Vid2World & H9
      & Official T10/H9 single-frame sliding rollout, repeated 32 times. \\
    \rowcolor{black!4}
    DIAMOND & H4
      & Official continuous four-frame-history rollout for 32 outputs. \\
    \midrule
    \rowcolor{black!8}
    \multicolumn{3}{c}{\textit{Embodied Navigation: RECON}} \\
    Vid2World & H4
      & Two consecutive 16-frame generations with predicted feedback. \\
    \rowcolor{black!4}
    NWM & H4
      & Official one-step four-frame sliding rollout, repeated 32 times. \\
    RAE-NWM & H4
      & Official latent one-step rollout, repeated 32 times before decoding. \\
    \bottomrule
  \end{tabular*}
\end{table}

For Vid2World, we retain the released domain-specific samplers rather than transplanting the RT-1 schedule to CSGO or RECON. WorldGym receives the released 10-D control representation obtained from the common raw 13-D RT-1 actions. DriftWorld generates on its native square canvas and is transformed to the common RT-1 scoring geometry only after prediction. NWM and RAE-NWM run at their released native resolutions; their decoded predictions are resampled once to the common RECON geometry. These transforms affect only the exported pixels, not the action encoder or the recurrent latent state.

\subsection{Frame Metrics and Statistical Aggregation}
\label{app:metric_definitions}

All RGB tensors are converted to $[0,1]$. For predicted frame $\widehat{\mathbf{o}}$ and target $\mathbf{o}$, MSE and MAE are means over all pixels and channels, and $\operatorname{PSNR}=10\log_{10}(1/(\operatorname{MSE}+10^{-8}))$. SSIM uses an $11\!\times\!11$ Gaussian window with standard deviation 1.5 and is evaluated on RGB channels. LPIPS uses the VGG backbone and inputs mapped to $[-1,1]$~\citep{zhang2018unreasonable}.

Metrics are computed independently for every predicted frame. We first average the 32 frame values within an episode and then take the arithmetic mean across episodes. The uncertainty after ``$\pm$'' is the sample standard deviation across episode-level means ($\mathrm{ddof}=1$), not a standard error and not variation across training seeds. PSNR is therefore a macro-average of frame-level PSNR, rather than PSNR obtained from a globally pooled MSE. The fourth main-table metric is MSE on RT-1 and MAE on CSGO and RECON, as indicated in Table~\ref{tab:main_results}.

Distributed evaluation partitions complete episodes only. Episode records are restored to canonical request order and aggregated in double precision on the CPU; no metric is reduced in a rank-dependent order. Deterministic kernels are enabled and TF32 is disabled. These choices make the reported statistics independent of the number of evaluation workers.

\subsection{Paired Uncertainty for the Main Table}
\label{app:main_table_paired}

\noindent\textbf{Statistical procedure.}
For every baseline in each domain--metric cell of Table~\ref{tab:main_results}, we compute the episode-wise paired difference on the same episodes, yielding 36 comparisons in total. Positive margins always favor \method{}: ours minus the baseline for PSNR and SSIM, and the baseline minus ours for LPIPS and MSE/MAE. We use 100,000 episode-level paired percentile-bootstrap resamples for the 95\% confidence interval and 100,000 two-sided paired sign-flip permutations for the $p$-value. All 36 comparisons form one family for Holm correction; a margin is counted as supported only when its confidence interval lies strictly above zero and its Holm-adjusted $p$-value is below 0.05.

\begin{table*}[t!]
  \centering
  \caption{\textbf{Paired uncertainty against every baseline.} Each cell reports the \method{}-favorable episode-mean margin $\Delta$ and its paired 95\% bootstrap interval. Positive values favor \method{}. All unmarked cells also pass the jointly Holm-corrected paired sign-flip test; $\dagger$ marks the sole statistically unresolved cell.}
  \label{tab:main_table_all_baselines_paired}
  \vspace{0.5em}
  \normalsize
  \setlength{\tabcolsep}{2pt}
  \setlength{\extrarowheight}{0pt}
  \renewcommand{\arraystretch}{1.0}
  \newcommand{\pairedcell}[3]{%
    \begin{tabular}[c]{@{}c@{}}%
      \strut $#1$\\[2pt]%
      \strut $[#2,\,#3]$%
    \end{tabular}}
  {\color{black}
  \begin{tabular*}{\textwidth}{@{}l@{\extracolsep{\fill}}cccc@{}}
    \toprule
    Baseline & PSNR $\uparrow$ & SSIM $\uparrow$
      & LPIPS $\downarrow$ & MSE/MAE $\downarrow$ \\[3pt]
    \midrule
    \rowcolor{black!8}
    \multicolumn{5}{c}{\rule[-1.0ex]{0pt}{3.2ex}%
      \textit{Robot Manipulation: RT-1 ($n{=}1{,}024$)}} \\[3pt]
    RLVR-World
      & \pairedcell{2.422}{2.330}{2.514}
      & \pairedcell{0.0592}{0.0574}{0.0609}
      & \pairedcell{0.0214}{0.0199}{0.0228}
      & \pairedcell{0.00355}{0.00333}{0.00377} \\[2pt]
    \rowcolor{black!4}
    Vid2World
      & \pairedcell{3.050}{2.947}{3.154}
      & \pairedcell{0.1144}{0.1115}{0.1173}
      & \pairedcell{0.0967}{0.0949}{0.0986}
      & \pairedcell{0.00501}{0.00475}{0.00529} \\[2pt]
    WorldGym
      & \pairedcell{2.650}{2.564}{2.736}
      & \pairedcell{0.0567}{0.0551}{0.0584}
      & \pairedcell{0.0577}{0.0564}{0.0591}
      & \pairedcell{0.00435}{0.00412}{0.00459} \\[2pt]
    \rowcolor{black!4}
    DriftWorld
      & \pairedcell{3.539}{3.443}{3.635}
      & \pairedcell{0.1079}{0.1053}{0.1104}
      & \pairedcell{0.1187}{0.1166}{0.1208}
      & \pairedcell{0.00650}{0.00622}{0.00678} \\[2pt]
    \midrule
    \rowcolor{black!8}
    \multicolumn{5}{c}{\rule[-1.0ex]{0pt}{3.2ex}%
      \textit{3D Game Simulation: CSGO ($n{=}70$)}} \\[3pt]
    DIAMOND
      & \pairedcell{1.317}{1.016}{1.626}
      & \pairedcell{0.0521}{0.0407}{0.0659}
      & \pairedcell{0.0601}{0.0508}{0.0698}
      & \pairedcell{0.02528}{0.01838}{0.03302} \\[2pt]
    \rowcolor{black!4}
    Vid2World
      & \pairedcell{1.675}{1.341}{2.014}
      & \pairedcell{0.0781}{0.0678}{0.0890}
      & \pairedcell{0.0028^{\dagger}}{-0.0065}{0.0119}
      & \pairedcell{0.02996}{0.02361}{0.03700} \\[2pt]
    \midrule
    \rowcolor{black!8}
    \multicolumn{5}{c}{\rule[-1.0ex]{0pt}{3.2ex}%
      \textit{Embodied Navigation: RECON ($n{=}128$)}} \\[3pt]
    NWM
      & \pairedcell{3.676}{3.469}{3.882}
      & \pairedcell{0.1359}{0.1293}{0.1427}
      & \pairedcell{0.1060}{0.0991}{0.1130}
      & \pairedcell{0.07285}{0.06811}{0.07762} \\[2pt]
    \rowcolor{black!4}
    RAE-NWM
      & \pairedcell{2.039}{1.834}{2.246}
      & \pairedcell{0.1223}{0.1176}{0.1271}
      & \pairedcell{0.0571}{0.0511}{0.0634}
      & \pairedcell{0.03907}{0.03584}{0.04233} \\[2pt]
    Vid2World
      & \pairedcell{1.107}{0.907}{1.311}
      & \pairedcell{0.0580}{0.0536}{0.0625}
      & \pairedcell{0.0146}{0.0079}{0.0214}
      & \pairedcell{0.02045}{0.01745}{0.02358} \\[2pt]
    \bottomrule
  \end{tabular*}}
\end{table*}

\noindent\textbf{Main-table findings.}
Table~\ref{tab:main_table_all_baselines_paired} shows that all 36 point
estimates favor \method{}, and 35 are supported by both analyses
after joint correction. This comprises all 16 RT-1 comparisons, all 12 RECON
comparisons, and seven of eight CSGO comparisons. The sole unresolved case is
CSGO LPIPS against Vid2World: its point estimate favors \method{} by $0.0028$,
but the paired interval $[-0.0065,0.0119]$ contains zero and
$p_{\mathrm{Holm}}=0.5599$. We therefore interpret this cell as statistically
unresolved rather than as a confirmed advantage. RT-1 is the primary domain
and is trained for 60k steps; the independently trained CSGO and RECON models
use a fixed 30k-step budget to test cross-domain applicability under limited
compute rather than exhaustive domain-specific tuning. Even under this
budget, 19 of their 20 baseline-specific margins are supported.

\subsection{Prediction Beyond the Training Horizon}
\label{app:rt1_h64}

\noindent\textbf{Long-horizon protocol and findings.}
To evaluate \method{} at temporal scales beyond those used for training, we
extend the RT-1 rollout from the maximum training horizon of AR32 to an
uninterrupted 64-frame prediction in Table~\ref{tab:rt1_h64_results}.
Constructing this H64 benchmark without
padding requires 64 aligned future RGB targets and the corresponding action
sequence after the common conditioning anchor. Exactly 35 of the 1,024
episodes in the frozen test protocol satisfy this requirement, and we evaluate
all 35 rather than sampling a subset. Eligibility is fixed solely by sequence
length before inference and is independent of any model output or metric. All
five methods are therefore evaluated on exactly the same ordered episodes,
common anchors, action sequences, and future targets. Each method retains its
native conditioning interface and continues a single rollout without being
restarted at step 32; every predicted frame in
$\mathbf{o}_{t+1:t+64}$ is scored using the main-table metrics and the same
episode-first aggregation. For \method{}, the recurrent state, action
sequence, and logical rollout step continue through all 64 predictions, while
steps 33--64 reuse the zero-based terminal trained time-conditioning index
($t{=}31$)
rather than querying embedding rows outside the AR32 training horizon.
\method{} ranks first on all four metrics.
Relative to the strongest baseline in each column, it improves PSNR by
$2.36$ dB and SSIM by $0.060$, while reducing LPIPS by $0.0243$ and MSE by
$0.00534$. Thus, its aggregate prediction-quality advantage persists at
twice the maximum training horizon. This evaluation establishes long-horizon
predictive quality; direct carrier correspondence, reappearance,
transport-route, and write-gate analyses are provided over all 35 episodes in
Table~\ref{tab:rt1_h64_persistence_mechanisms} and visualized for an
occlusion--reappearance example in
Figure~\ref{fig:carrier_transport_write}.

\begin{table*}[t!]
  \centering
  \caption{\textbf{64-step RT-1 prediction.} Mean $\pm$ episode sample standard deviation over all 35 of the 1,024 frozen test episodes that contain 64 aligned future frames. Metrics average $\mathbf{o}_{t+1:t+64}$; the common anchor $\mathbf{o}_{t}$ and any method-native history are conditioning only. Best and second-best results are shown in \textbf{bold} and \underline{underlined}.}
  \label{tab:rt1_h64_results}
  \vspace{0.5em}
  \normalsize
  \setlength{\tabcolsep}{1pt}
  \renewcommand{\arraystretch}{1.12}
  {\color{black}
  \begin{tabular*}{\textwidth}{@{}l@{\extracolsep{\fill}}cccc@{}}
    \toprule
    \rowcolor{black!8}
    \multicolumn{5}{c}{\textit{Robot Manipulation: RT-1 (64 Future Frames)}} \\
    \midrule
    Model & PSNR $\uparrow$ & SSIM $\uparrow$ & LPIPS $\downarrow$ & MSE $\downarrow$ \\
    \midrule
    RLVR-World
      & $\underline{20.22 \pm 2.78}$
      & $0.730 \pm 0.099$
      & $\underline{0.213 \pm 0.075}$
      & $\underline{0.0138 \pm 0.0096}$ \\
    \rowcolor{black!4}
    Vid2World
      & $19.04 \pm 2.35$
      & $0.636 \pm 0.107$
      & $0.315 \pm 0.067$
      & $0.0175 \pm 0.0116$ \\
    WorldGym
      & $19.98 \pm 2.31$
      & $\underline{0.736 \pm 0.093}$
      & $0.250 \pm 0.067$
      & $0.0144 \pm 0.0074$ \\
    \rowcolor{black!4}
    DriftWorld
      & $19.51 \pm 2.38$
      & $0.678 \pm 0.094$
      & $0.325 \pm 0.068$
      & $0.0154 \pm 0.0083$ \\
    \rowcolor{cyan!6}
    \textbf{\method{} (Ours)}
      & $\bm{22.58 \pm 2.80}$
      & $\bm{0.795 \pm 0.095}$
      & $\bm{0.188 \pm 0.076}$
      & $\bm{0.0085 \pm 0.0071}$ \\
    \bottomrule
  \end{tabular*}}
  \vspace{-8pt}
\end{table*}

\paragraph{Persistent-state mechanism diagnostics.}

To determine whether the recurrent state operates as intended rather than
merely improving next-frame reconstruction, we analyze the internal mechanism
over the same complete set of 35 H64 episodes. At every predicted step, we
query both the current carrier descriptors and their canonical-content
descriptors against all 16 slots in the bank initialized from the Context
frame. Let $\mathbf{c}_i^t$ denote carrier $i$'s dynamic descriptor and
$\mathbf{q}_i^t$ its canonical-content descriptor after predicting frame $t$;
$t=0$ denotes the Context-initialized bank. For either descriptor family
$\mathbf{z}\in\{\mathbf{c},\mathbf{q}\}$ and slots
$i,j\in\{1,\ldots,16\}$, define
\begin{equation}
\begin{aligned}
C_{ij}^t(\mathbf{z})&=\operatorname{cos}(\mathbf{z}_i^t,\mathbf{z}_j^0),
&\qquad \mathrm{diag}_t(\mathbf{z})&=\frac{1}{16}\sum_i C_{ii}^t(\mathbf{z}),\\
\mathrm{ID}_t(\mathbf{z})&=\frac{1}{16}
\left|\left\{i:\operatorname*{arg\,max}_j C_{ij}^t(\mathbf{z})=i\right\}\right|,
&\qquad \Delta_t(\mathbf{z})&=\frac{1}{16}\sum_i\left(
C_{ii}^t(\mathbf{z})-\max_{j\ne i}C_{ij}^t(\mathbf{z})\right).
\end{aligned}
\label{eq:app_carrier_correspondence}
\end{equation}
Here $\operatorname{cos}$ is cosine similarity and $|\cdot|$ denotes set
cardinality. In Eq.~\plaineqref{eq:app_carrier_correspondence},
$\mathrm{diag}_t$ is mean same-slot similarity,
$\mathrm{ID}_t$ is \emph{same-slot retrieval}---the fraction for which the
corresponding Context slot is the nearest neighbor---and the
\emph{separation margin} subtracts the strongest other-slot cosine from the
same-slot cosine, thereby distinguishing a uniquely traceable identity from a
bank whose slots have merely become mutually similar. Reporting cosine,
retrieval, and margin together is important: retrieval alone could reflect the
fixed slot organization, whereas a positive margin and high canonical-content
similarity show that each slot remains distinctly associated with its original
stored content.

\noindent\textbf{Transport and write statistics.}
We measure transport using the actual per-cell continuous Lagrangian mixture
coefficient $m\in[0,1]$ in Eq.~\plaineqref{eq:app_hybrid_flow}, rather than
interpreting transport as a discrete token route. Here $\mathbf{f}^{E}$ is
dense Eulerian flow, $\mathbf{f}^{L}$ is carrier-derived Lagrangian flow, and
$\mathbf{f}$ is the committed flow. We report both the mean of $m$ and the
fraction of spatial locations at which the carrier-derived component is the
majority ($m>0.5$); neither quantity is a motion magnitude. We then quantify
revision after carrying by the effective interpolation coefficient $w$ in
$\mathbf{x}^{+}=(1-w)\mathbf{x}_{\mathrm{carry}}+
w\mathbf{x}_{\mathrm{new}}$. Thus $w=0$ retains carried state and $w=1$
replaces it. Scene, carrier, canonical-feature, and canonical-alpha writes
all use this direction; the canonical coefficients include their evidence and
confidence modulation and are not separate biased gates.
Every statistic is first reduced over the indicated steps and internal slots
or spatial locations within an episode. Scene-write $q_{0.90}$ is the 90th
percentile of all scene-write coefficients over those same steps and spatial
cells within one episode. Table~\ref{tab:rt1_h64_persistence_mechanisms}
then reports the mean and sample standard deviation of the resulting 35
episode-level values. The Steps 33--64 and $t{=}64$ columns therefore probe
state evolution strictly beyond the AR32 training horizon.

\begin{table*}[t!]
  \centering
  \caption{\textbf{Persistent-state mechanism diagnostics over 64-step RT-1 rollouts.} Mean $\pm$ episode sample standard deviation over all 35 eligible episodes. Each statistic is reduced within an episode before cross-episode aggregation; percentages display shares and gate values multiplied by 100. These internal diagnostics are descriptive and do not share a universal better direction.}
  \label{tab:rt1_h64_persistence_mechanisms}
  \vspace{0.5em}
  \normalsize
  \setlength{\tabcolsep}{5pt}
  \renewcommand{\arraystretch}{1.10}
  {\color{black}
  \begin{tabular*}{\textwidth}{@{\extracolsep{\fill}}lccc@{}}
    \toprule
    Diagnostic
      & Steps 1--32
      & Steps 33--64
      & $t{=}64$ \\
    \midrule
    \rowcolor{black!8}
    \multicolumn{4}{c}{\textit{Carrier identity relative to Context}} \\
    Same-slot retrieval (\%)
      & $100.0 \pm 0.0$ & $100.0 \pm 0.0$ & $100.0 \pm 0.0$ \\
    \rowcolor{black!4}
    Same-slot cosine
      & $0.924 \pm 0.002$ & $0.753 \pm 0.003$ & $0.682 \pm 0.003$ \\
    Separation margin
      & $0.466 \pm 0.003$ & $0.380 \pm 0.004$ & $0.344 \pm 0.006$ \\
    \midrule
    \rowcolor{black!8}
    \multicolumn{4}{c}{\textit{Canonical content relative to Context}} \\
    Same-slot retrieval (\%)
      & $100.0 \pm 0.0$ & $100.0 \pm 0.0$ & $100.0 \pm 0.0$ \\
    \rowcolor{black!4}
    Same-slot cosine
      & $0.993 \pm 0.002$ & $0.980 \pm 0.006$ & $0.977 \pm 0.006$ \\
    Separation margin
      & $0.740 \pm 0.069$ & $0.728 \pm 0.069$ & $0.726 \pm 0.069$ \\
    \midrule
    \rowcolor{black!8}
    \multicolumn{4}{c}{\textit{Transport-route usage}} \\
    Mean Lagrangian flow share, $m$ (\%)
      & $85.0 \pm 2.1$ & $76.3 \pm 4.1$ & $69.4 \pm 5.6$ \\
    \rowcolor{black!4}
    Locations with $m>0.5$ (\%)
      & $99.2 \pm 1.2$ & $92.8 \pm 3.7$ & $82.5 \pm 8.5$ \\
    \midrule
    \rowcolor{black!8}
    \multicolumn{4}{c}{\textit{Post-carry write gates}} \\
    Scene write, mean (\%)
      & $2.64 \pm 0.51$ & $1.84 \pm 0.46$ & $1.76 \pm 0.52$ \\
    \rowcolor{black!4}
    Scene write, $q_{0.90}$ (\%)
      & $6.05 \pm 0.90$ & $4.54 \pm 0.96$ & $4.34 \pm 1.19$ \\
    Carrier write, mean (\%)
      & $0.46 \pm 0.07$ & $0.54 \pm 0.18$ & $0.58 \pm 0.20$ \\
    \rowcolor{black!4}
    Canonical-feature write, mean (\%)
      & $0.97 \pm 0.03$ & $0.18 \pm 0.02$ & $0.11 \pm 0.02$ \\
    Canonical-alpha write, mean (\%)
      & $1.44 \pm 0.02$ & $0.27 \pm 0.02$ & $0.17 \pm 0.03$ \\
    \bottomrule
  \end{tabular*}}
  \vspace{-6pt}
\end{table*}

\noindent\textbf{Carrier identity and content persistence.}
The resulting pattern separates evolving motion state from persistent content.
Carrier-to-Context cosine decreases from $0.924$ over Steps 1--32 to $0.682$
at $t{=}64$, showing that the carrier descriptors adapt throughout the rollout
rather than remaining artificially frozen. Nevertheless, the corresponding
Context slot is the nearest neighbor for every carrier at every horizon, and
the separation margin remains clearly positive at $t{=}64$
($0.344 \pm 0.006$). Carrier identity therefore remains uniquely traceable
even as its dynamic descriptor evolves. More importantly, the canonical
content bound to those carriers is substantially more stable: its same-slot
cosine is $0.977 \pm 0.006$ at $t{=}64$, its retrieval remains $100\%$, and
its separation margin changes only from $0.740$ to $0.726$. This separation
between an evolving carrier descriptor and nearly invariant carried content
is the behavior targeted by the persistent Lagrangian representation.

\noindent\textbf{Transport and selective writing.}
The forward transition also continues to use that representation beyond the
training horizon. Mean Lagrangian flow share is $76.3\%$ over Steps 33--64
and $69.4\%$ at $t{=}64$; carrier flow is the majority component at $92.8\%$
of late-horizon spatial locations and $82.5\%$ of locations at the terminal
step. These are continuous mixture shares, rather than counts of a discrete
route, and show that the dense Eulerian fallback remains complementary without
replacing carrier transport. Finally, the recurrent state is not rewritten
wholesale after carrying. At $t{=}64$, the post-carry scene-write gate
averages $1.76\%$, while the cross-episode mean of its within-episode 90th
percentile is only $4.34\%$; carrier, canonical-feature, and canonical-alpha
writes average
$0.58\%$, $0.11\%$, and $0.17\%$, respectively. The small but nonzero gates
permit localized innovations while retaining transported evidence as the
default state-update path. Together with the H64 output advantage in
Table~\ref{tab:rt1_h64_results} and the occlusion--reappearance example in
Figure~\ref{fig:carrier_transport_write}, these results support the intended
carry-before-write mechanism: persistent content remains bound to uniquely
traceable carriers, is transported through the recurrent state, and is
updated selectively rather than regenerated at every prediction step.

\subsection{Motion-Fidelity Diagnostics}
\label{app:motion_metrics}

The three diagnostics in Figure~\ref{fig:rt1_motion} use one frozen
RAFT estimator for every predicted and reference rollout
\citep{teed2020raft}. Let $\widehat{\mathbf{u}}_{\tau}$ and
$\mathbf{u}^{\star}_{\tau}$ be flows estimated from adjacent predicted and
reference frames. Flow EPE averages
$\|\widehat{\mathbf{u}}_{\tau}-\mathbf{u}^{\star}_{\tau}\|_2$ over pixels
and transitions. It is a discrepancy between two estimated flows, not an
error against measured flow annotations. Flow Cosine evaluates their
directional agreement only where the reference forward-flow magnitude is at
least one pixel. Dynamic-region LPIPS instead uses reference backward flow
to define a target-coordinate motion mask with the same threshold. Both
masks are fixed by the reference video and shared by all methods. Aggregation
again proceeds within episode and then across the same 1,024 episodes.

These quantities are protocol-specific diagnostics designed to reduce the
dominance of static background. They are not substituted for the main
appearance metrics, and neither their threshold nor their reported transition
range is selected separately for an individual method.

\section{Ablation Protocol and Paired Analysis}
\label{app:ablation_details}

\noindent\textbf{Controlled training protocol.}
All variants in Table~\ref{tab:rt1_ablation_30k} are independently trained
from scratch rather than obtained by disabling pathways in a trained
checkpoint. The four RT-1 seed-0 models are all 0.8B-scale and use the same
tokenizer, training episodes, action representation, AdamW schedule, BF16
precision, full BPTT without teacher forcing,
AR8$\rightarrow$AR16$\rightarrow$AR32 curriculum, global batch of 32 on the
same $4\!\times\!8$-GPU topology, 30k-update budget, and ordered
1,024-episode H32 evaluation protocol. Their active parameter counts are
824.450M (Plain AR), 819.397M (Recurrent Memory), 808.397M (Dense-only), and
818.746M (Full). Dense-only removes the
carrier-specific machinery without introducing an auxiliary capacity-matching
module, thereby preserving a clean structural ablation rather than adding a
new architectural component.
The independent-seed RT-1 pair and the CSGO pair retain the matched recipe
within their respective settings.

\noindent\textbf{State and objective matching.}
Full and Dense-only share the 320-token dense Eulerian scene, 64 generic
interaction slots, and four-frame hard-prediction history. Dense-only retains
the applicable innovate/stationary route, slot-diversity, and dense-flow terms,
but structurally removes carriers, canonical content, source pointers,
transported routing, and all losses defined on them. Recurrent Memory instead
uses the latest hard prediction and an $80\!\times\!1280$ action-conditioned
GRU state, while Plain AR retains only the latest hard prediction; neither has
a scene, slot, route, or carrier variable. All four retain the same applicable
joint-code, generation-code, digit, ordered-value, and frozen-decoder targets,
with identical coefficients and schedules. Terms whose state variable does
not exist are omitted rather than replaced by a variant-specific auxiliary
objective. Thus the histories are intentionally architecture-specific, while
the data, prediction targets, optimization, sample/update budget, and 0.8B
model scale remain controlled. Table~\ref{tab:app_ablation_contract} summarizes these
state interventions, and Table~\ref{tab:app_complete_loss} gives the exact Full
and Plain-AR loss correspondence.

\begin{table}[t!]
  \centering
  \color{black}
  \caption{\textbf{Scope of the trained RT-1 ablations.} Every variant is trained independently from scratch for 30k steps and evaluated on the same ordered 1,024-episode protocol.}
  \label{tab:app_ablation_contract}
  \vspace{0.5em}
  \normalsize
  \setlength{\tabcolsep}{1.5pt}
  \renewcommand{\arraystretch}{1.13}
  \begin{tabular*}{\linewidth}{@{\extracolsep{\fill}}>{\raggedright\arraybackslash}m{0.18\linewidth}>{\raggedright\arraybackslash}m{0.24\linewidth}>{\raggedright\arraybackslash}m{0.50\linewidth}@{}}
    \toprule
    \rowcolor{black!8}
    \multicolumn{3}{c}{\textit{RT-1, seed 0 (matched 30k)}} \\
    \midrule
    Variant & Recurrent state & Controlled role \\
    \midrule
    Plain AR
      & Latest hard-predicted Dynamic state
      & Removes Inertial Memory and directly predicts the next Dynamic state
        from fixed Context, the action, rollout step, and preceding
        prediction. \\
    \rowcolor{black!4}
    Recurrent Memory
      & Latest hard prediction $+$ $80\!\times\!1280$ GRU
      & Tests generic learned recurrence without the structured
        Eulerian--Lagrangian state or persistent carrier transport. \\
    Dense-only
      & 320-token scene $+$ 64 slots $+$ four-frame history
      & Retains the dense state and matched update recipe while removing
        regional carriers, canonical content, and carrier transport. \\
    \rowcolor{cyan!6}
    \textbf{Full}
      & Dense-only state $+$ 16 Lagrangian carriers
      & Executes the complete carry--predict--update transition and therefore
        measures the incremental effect of carriers relative to Dense-only. \\
    \bottomrule
  \end{tabular*}
\end{table}

\noindent\textbf{Track and reappearance selection.}
Track and reappearance selection is fixed by ground-truth DINO ViT-B/16
features; predictions never determine which locations are scored. A Context
patch is eligible when its cosine separation from the closest other Context
patch is at least $0.02$. At a later frame, a reliable correspondence must be
a mutual nearest neighbor with cosine at least $0.60$, separation margin at
least $0.02$, and at least eight reliable observations accumulated along that
track.
For each scored destination patch, let $\widehat{\mathbf{z}}$ and
$\mathbf{z}^{\star}$ be the $\ell_2$-normalized DINO features from the
predicted frame and its corresponding ground-truth frame. The content error is
\begin{equation}
  e(\widehat{\mathbf{z}},\mathbf{z}^{\star})
  =1-\widehat{\mathbf{z}}^{\mathsf T}\mathbf{z}^{\star}.
  \label{eq:app_dino_content_error}
\end{equation}
Thus lower values indicate that predicted content at the ground-truth-defined
destination retains the target appearance in feature space.
For the track-based metrics, the H32 rollout is divided at its fixed midpoint:
Steps 1--16 form the early half and Steps 17--32 the late half. The primary
long-lived and late-track measures use the late half to emphasize accumulated
content persistence rather than short-range prediction quality; each is the
episode mean of Eq.~\plaineqref{eq:app_dino_content_error} over its eligible
late-half track points. This boundary is identical for every method, seed,
and domain. Post-reappearance content is the episode mean over all qualifying
post-gap destinations anywhere in H32 and is not restricted to the late-half
window.
Long-lived points and post-reappearance events additionally require
displacement of at least two DINO patches. A reappearance event requires at
least four intervening absent steps and two consecutive confirmed observations
immediately before and after the gap. These events are an automatic
ground-truth feature-reappearance proxy, not manually annotated physical
occlusions.

\begin{table*}[t!]
  \centering
  \color{black}
  \caption{\textbf{Paired persistence improvements.} Each baseline cell reports the Full-model-favorable episode-mean difference $\Delta=\text{baseline}-\text{Full}$ and its pointwise paired 95\% percentile-bootstrap interval obtained from 50,000 bootstrap replicates of the same eligible paired episodes, not from additional test episodes. Positive values favor the Full model; blue rows identify the reference method.}
  \label{tab:app_ablation_paired}
  \vspace{0.2em}
  \normalsize
  \setlength{\tabcolsep}{1.5pt}
  \renewcommand{\arraystretch}{1.06}
  \newcommand{\persistcell}[3]{$#1\,[#2,#3]$}
  \begin{tabular*}{\textwidth}{@{}l@{\extracolsep{\fill}}ccc@{}}
    \toprule
    \rowcolor{black!8}
    \multicolumn{4}{c}{\textit{RT-1, seed 0}} \\
    \midrule
    Model
      & \shortstack{Long-lived\\content}
      & \shortstack{Post-reappearance\\content}
      & \shortstack{Late-track\\content} \\
    \midrule
    Plain AR
      & \persistcell{0.0999}{0.0932}{0.1067}
      & \persistcell{0.1108}{0.1007}{0.1210}
      & \persistcell{0.0133}{0.0121}{0.0145} \\
    \rowcolor{black!4}
    Recurrent Memory
      & \persistcell{0.0269}{0.0218}{0.0321}
      & \persistcell{0.0339}{0.0260}{0.0419}
      & \persistcell{0.0059}{0.0051}{0.0067} \\
    Dense-only
      & \persistcell{0.0102}{0.0054}{0.0149}
      & \persistcell{0.0147}{0.0072}{0.0222}
      & \persistcell{0.0020}{0.0010}{0.0032} \\
    \rowcolor{cyan!6}
    \textbf{\method{} (Full)}
      & \multicolumn{3}{c}{\textbf{Reference method}} \\
    \midrule
    \rowcolor{black!8}
    \multicolumn{4}{c}{\textit{RT-1, seed 1}} \\
    \midrule
    Plain AR
      & \persistcell{0.1019}{0.0952}{0.1086}
      & \persistcell{0.1020}{0.0920}{0.1121}
      & \persistcell{0.0163}{0.0148}{0.0177} \\
    \rowcolor{cyan!6}
    \textbf{\method{} (Full)}
      & \multicolumn{3}{c}{\textbf{Reference method}} \\
    \midrule
    \rowcolor{black!8}
    \multicolumn{4}{c}{\textit{CSGO, seed 0}} \\
    \midrule
    Plain AR
      & \persistcell{0.0396}{0.0343}{0.0450}
      & \persistcell{0.0337}{0.0274}{0.0401}
      & \persistcell{0.0379}{0.0331}{0.0427} \\
    \rowcolor{cyan!6}
    \textbf{\method{} (Full)}
      & \multicolumn{3}{c}{\textbf{Reference method}} \\
    \bottomrule
  \end{tabular*}
  \vspace{-4pt}
\end{table*}

\noindent\textbf{Episode support and aggregation.}
On RT-1, the long-lived, post-reappearance, and late-track statistics contain
994, 843, and 1,024 paired episodes, respectively; the reappearance set
contains 3,123 events from 843 episodes. The corresponding CSGO counts are
359, 349, and 369 episodes, with 3,849 events from 349 episodes. We first
average patches or events within an episode and then aggregate episodes, so a
single rollout with many tracks cannot dominate the reported uncertainty.

\noindent\textbf{Paired findings.}
Every pointwise interval in Table~\ref{tab:app_ablation_paired} lies strictly above
zero. Most importantly, the Full--Dense-only comparison isolates the
incremental contribution of regional carriers after both architectures are
trained from scratch: the carrier-equipped Full model reduces long-lived,
post-reappearance, and late-track errors on the paired evaluation. The larger Full
advantages over Recurrent Memory and Plain AR establish that the complete
structured state is preferable to generic recurrence or direct
autoregression. Finally, the independent RT-1 seed and CSGO comparisons
reproduce the Full--Plain advantage on all three measures, showing that the
result is not specific to one training run or interaction domain.